# CAN-PINN: A Fast Physics-Informed Neural Network Based on Coupled-Automatic-Numerical Differentiation Method

Pao-Hsiung Chiu[a,†], Jian Cheng Wong[a,b,*,†], Chinchun Ooi[a], My Ha Dao[a], Yew-Soon Ong[a,b]

[a] *Agency for Science, Technology and Research (A*STAR), 138632, Singapore*
[b] *School of Computer Science and Engineering, Nanyang Technological University, 639798, Singapore*

**Abstract** – In this study, novel physics-informed neural network (PINN) methods for coupling neighboring support points and their derivative terms which are obtained by automatic differentiation (AD), are proposed to allow efficient training with improved accuracy. PINNs constrain their training loss function with ordinary and partial differential equations, to ensure outputs obey the governing physics. The computation of differential operators required for loss evaluation at collocation points are conventionally obtained via automatic differentiation. Although AD method has the advantage of being able to compute the exact gradients at any point, such PINNs can only achieve high accuracies with large numbers of collocation points—otherwise they are prone to optimizing towards unphysical solution. To make PINN training fast, the dual ideas of using numerical differentiation (ND)-inspired method and coupling it with AD are employed to define the loss function. The ND-based formulation for training loss can strongly link neighboring collocation points to enable efficient training in sparse sample regimes, but its accuracy is restricted by the interpolation scheme. The proposed *coupled-automatic-numerical* differentiation framework—labeled as *can*-PINN—unifies the advantages of AD and ND, providing more robust and efficient training than AD-based PINNs, while further improving accuracy by up to 1-2 orders of magnitude relative to ND-based PINNs. For a proof-of-concept demonstration of this *can*-scheme to fluid dynamic problems, two numerical-inspired instantiations of *can*-PINN schemes for the convection and pressure gradient terms were derived to solve the incompressible Navier-Stokes (N-S) equations. Theoretical analysis shows that the proposed *can*-schemes have smaller dispersion and dissipation errors than the baseline ND-based schemes. The superior performance of *can*-PINNs is demonstrated on several challenging problems, including the flow mixing phenomena, lid driven flow in a cavity, and channel flow over a backward facing step. The results reveal that for challenging problems like these, *can*-PINNs can consistently achieve very good accuracy whereas conventional AD-based PINNs fail.

## 1. Introduction

Physics-informed machine learning [1], in particular physics-informed neural networks (PINNs)—as per Raissi *et al.* [2]— have received increasing attention in recent years. PINNs leverage the expressiveness of deep neural networks (DNNs) to model the dynamical evolution $\hat{u}(x,t;w)$ of physical systems in space $x \in \Omega$ and time $t \in [0,T]$ via the optimization of network parameters $w$. The central idea of PINNs is to incorporate the governing laws of such systems, typically in the form of ordinary differential equations (ODEs) or partial differential equations (PDEs), into the training loss function. The PINN training then aims to reduce the residual loss of the differential equations for the model output $\hat{u}(x,t;w)$, over a set of collocation points $(x,t)$ sampled from the problem domain $\Omega \times [0,T]$. This physics-informed loss function constrains the PINN from violating the differential equations and the prescribed initial conditions (ICs) and boundary conditions (BCs), ensuring that its output obeys the governing physics given only limited or even zero labelled data. For the latter case, PINNs essentially form a new class of mesh-free methods to solve differential equations,in which the problem is transformed into a neural network optimization [3].

---

[*] Corresponding author.
   *Email address:* wongj@ihpc.a-star.edu.sg (Jian Cheng Wong)
[†] Equal contribution.

The concept of PINNs can be traced back to 1990s, where neural algorithms for solving differential equations were proposed [4–8]. Since then, both neural network methodologies and compute capability have greatly progressed. Fueled by advances in deep learning, a variety of PINN models and applications have been proposed in the past few years. Different architectures, from fully connected neural networks [2,3,9–12] to convolutional neural networks [13–16], to recurrent neural networks [17–20], and generative adversarial neural networks [21,22], have been explored in the context of PINN. PINNs have been demonstrated for various physics phenomena, including heat transfer [23,24], fluid dynamics [25–27] and electromagnetic propagation [28–30]. Moreover, PINNs offer advantage of being seamlessly extended to tackle real world inverse problems [2,31]. Interesting applications include quantification of cardiovascular flows from visualization or sensor data [32–34], metamaterials design [35], and nondestructive quantification of cracks [36].

Despite the potential for a wide range of physic phenomena and applications, training an accurate PINN model remains a challenge [1]. There have been significant efforts of late to improve PINN trainability, which include learning in sinusoidal spaces [37,38], adaptively calibrating the composition of loss components during training [39–42], and importance sampling [43], to name just a few. Nevertheless, today's PINNs are still computationally demanding. In general, a huge amount of collocation points is required for matching the differential equations in order to train a good PINN model [42,43]. The PINN model then needs to be trained with a large number of optimization iterations [44,45]. The amount of collocation points and training iterations tend to increase with the problem complexity in practice, further complicating the already non-trivial task of finding appropriate hyper-parameters for effective training.

The vast majority of recent PINN implementations favor the fully connected DNN architecture [46–48], where the computation of differential operators—required for evaluating the differential equations' residual loss—at collocation points can be conveniently obtained via AD [49] during training. For training these PINN models, one can also numerically compute the differential operators, such as via the central difference or finite volume method [50]. Ren *et al.* [20] employed the finite-difference-based filters to their Physics-informed convolutional-recurrent network. Wandel *et al.* [14,15] also proposed to employ convolutional neural network and evaluate the loss functions by finite difference method based on a Marker-And-Cell (MAC) grid. Gao *et al.* [16] proposed the Physics-informed geometry-adaptive convolutional neural networks for solving PDEs on irregular domain. They also employed the finite-difference-based filters when evaluating the loss function. However, coordinate transformations need to be performed to handle the irregular domain. To naturally handle the complex geometries, Gao *et al.* [51] further proposed the Physics-informed graph neural Galerkin networks, which utilized nodal continuous Galerkin method. Alternatively, Haghighat *et al.* [52] utilized the peridynamic differential operator to develop a non-local PINN for solving PDEs. The differential operators computed by ND and AD are very different in nature and they have own merits in PINN implementation. For example, while ND approximates the gradients from a local set of PINN outputs based on certain numerical scheme, AD has the advantage of being able to compute the exact gradients at any point.

In the present study, we show that PINNs with training loss computed by AD (referred to as *a*-PINNs) can only be accurately trained with huge amount of collocation points. The *a*-PINN training becomes completely unrelated to their accuracy with insufficient collocation points, i.e., even when the training losses have been optimized to a very small value, *a*-PINNs can still be far from the true solution. Therefore, we propose to compute PINNs training loss by ND (referred to as *n*-PINNs) based on the sampling points and a finite difference-type stencil without requiring a predefined mesh topology. The proposed *n*-PINNs are more robust to the amount of collocation points, and capable of efficiently approximating the right solution with much less collocation points than required by *a*-PINNs. However, they may be less accurate than *a*-PINNs given large quantities of collocation points in some cases, depending on the accuracy of the numerical scheme.

Given the respective observed advantages and disadvantages of *a*-PINNs and *n*-PINNs, we further propose a novel *coupled-automatic-numerical* differentiation scheme for computing the PINN training loss, dubbed as *can*-PINNs. *can*-PINNs inherit the merits of both *a*-PINNs and *n*-PINNs, in that it robustly and efficiently produces accurate solutions even with minimal collocation points during training, unlike *a*-PINNs, and yet is more accurate than *n*-PINNs. As an illustration of the proposed methodology, we derive two versions of *can*-PINN based on the upwind and central difference numerical schemes commonly employed for differential operators. Note that the formulation of *can*-PINN presented in this work is generic and can be extended to other schemes which are based on Taylor series expansion of varying form and accuracy. We then carry out fundamental analysis of the two proposed *can*-PINN schemes, showing that they are more accurate than baseline *n*-PINNs, and validate these methods on a synthetic ODE problem. The superior performance of *n*-PINNs and *can*-PINNs are then demonstrated on several challenging PINN problems, i.e., predicting without any labelled data 1) the flow mixing phenomena governed by pure convection equation, 2) the lid driven flow in cavity and 3) the channel flow over a backward facing step, which are both governed by incompressible N-S equations. In addition, we demonstrate the efficacy of *can*-PINNs on inverse problem for N-S equations, being able to correctly infer the unknown Reynolds number based on very sparse observations. Our extensive experiments show that the proposed *can*-PINN is indeed highly efficient, allowing us to tackle challenging differential equation problems where *a*-PINNs fail, while consistently providing more accurate solutions than *n*-PINNs.

The remainder of the paper is organized as follows. In Section 2.1 to 2.3, we describe the PINNs with training loss computed by AD, ND and the proposed *coupled-automatic-numerical* differentiation. Fundamental analysis for the proposed method is covered in Section 2.4. Section 3 enumerates the extensive experimental studies conducted to illustrate the relative advantages of *n*-PINNs and *can*-PINNs across multiple forward and inverse modelling problems. Concluding remarks and direction for future research are then presented in Section 4.

## 2. Methodology

*2.1. Overview of PINNs with automatic differentiation (a-PINNs)*

In this section, we briefly outline the PINN methodology as commonly employed in current literature [2] and software [46–48]. A typical PINN uses a fully connected DNN architecture to represent the solution of the dynamical process $u$. The PINN model predicts $\hat{u}(x, t; \boldsymbol{w})$ given the spatial $x \in \Omega$ and temporal $t \in [0, T]$ inputs. The spatial domain usually has 1-, 2- or 3-dimensions in most physical problems. The accuracy of the PINN outputs is determined by the network parameters $\boldsymbol{w}$, which are optimized w.r.t. the PINN loss function during the training. To derive the PINN loss function, we consider $u$ to be mathematically described by differential equations of the general form:

$$\mathcal{N}_t[u(x,t)] + \mathcal{N}_x[u(x,t)] = 0, \quad x \epsilon \Omega, t \epsilon (0,T], \quad \text{(1a)}$$
$$u(x, 0) = u_o(x), \quad x \epsilon \Omega, \quad \text{(1b)}$$
$$\mathcal{B}[u(x,t)] = g(x,t), \quad x \epsilon \partial\Omega, t \epsilon (0,T], \quad \text{(1c)}$$

where $\mathcal{N}_t[\cdot]$ and $\mathcal{N}_x[\cdot]$ are the general differential operator which can include any combination of linear and non-linear terms of temporal and spatial derivatives, such as the time derivative, the first and second order spatial derivatives $u_t(x,t)$, $u_x(x,t)$ and $u_{xx}(x,t)$, respectively. The initial condition at $t = 0$ is defined by $u_o(x)$. The boundary operator $\mathcal{B}[\cdot]$, which can be an identity operator (Dirichlet boundary condition), a differential operator (Neumann boundary condition) or a mixed identity-differential operator (Robin boundary condition), enforces the desired condition $g(x, t)$ at the domain boundary $\partial\Omega$.

Then, the PINN training loss function is defined as:

$$\mathcal{L} = \mathcal{L}_{Data} + \lambda_{DE}\mathcal{L}_{DE} + \lambda_{IC}\mathcal{L}_{IC} + \lambda_{BC}\mathcal{L}_{BC}, \quad \text{(2a)}$$

which includes the data loss component when data is available, e.g., under inverse problem scenario,

$$\mathcal{L}_{Data} = \frac{1}{n}\sum_{i=1}^{n}(u_i - \hat{u}_i)^2, \quad \text{(2b)}$$

and the PDE loss components,

$$\mathcal{L}_{DE} = \|\hat{u}_t(\cdot\,;\boldsymbol{w}) + \mathcal{N}_x[\hat{u}(\cdot\,;\boldsymbol{w})]\|^2_{\Omega \times (0,T]}, \quad \text{(2c)}$$
$$\mathcal{L}_{IC} = \|\hat{u}(\cdot\,,0\,;\boldsymbol{w}) - u_0\|^2_{\Omega}, \quad \text{(2d)}$$
$$\mathcal{L}_{BC} = \|\mathcal{B}[\hat{u}(\cdot\,;\boldsymbol{w})] - g(\cdot)\|^2_{\partial\Omega \times (0,T]}. \quad \text{(2e)}$$

The relative weights, $\lambda$s in (2a), control the trade-off between different components in the loss function. The right scaling significantly speeds up the convergence rate of PINN training [39,53]. Hence, it is important to use an appropriate scaling strategy depending on the problem at hand. The computation of the loss described by (2) involves matching the PINN output $\hat{u}$ against target $u$ over $n$ labelled samples (2b), substitution of the output $\hat{u}$ into the differential equations for evaluating the residuals over the problem domain $\Omega \times (0,T]$ (2c), as well as matching the output $\hat{u}$ against initial conditions at $t = 0$ over the problem domain $\Omega$ (2d), and boundary conditions over the domain boundary $\partial\Omega$ and time $(0,T]$ (2e). When solving a forward differential equation problem, the data loss component $\mathcal{L}_{Data}$ (2b) is omitted.

The PDE loss components (2c-e) are defined over a continuous domain, but for practical reasons, we compute the residuals over a finite set of $m$ collocation points $D = \{(x_i, t_i)\}_{i=1}^{m}$ during training. These collocation points are sampled from the problem domain, for example, using an equidistantly spaced grid, randomized Latin hypercube sampling or importance sampling strategy. Differential operators, such $\hat{u}_t(x,t;\boldsymbol{w})$, $\hat{u}_x(x,t;\boldsymbol{w})$, $\hat{u}_{xx}(x,t;\boldsymbol{w})$, are required for the evaluation of the residuals in PDE loss on these collocation points. When the PINN with fully connected DNN architecture is higher order differentiable w.r.t. its inputs $(x,t)$—given that the activation function is higher order differentiable—the computation of differential operators can then be conveniently obtained via AD—which is already in place for computing the gradients of $\boldsymbol{w}$ for the optimization—at any collocation point. This makes AD the default method for computing the training loss for PINNs (*a*-PINNs). Finally, state-of-the-art algorithms such as ADAM [54] are used for optimizing the PINN weights $\boldsymbol{w}$.

Note that although a "tanh" activation function is most widely used for PINNs, it was suggested by recent studies that learning in the sinusoidal space of PINNs can achieve a more accurate solution [37,38]. In present study, we adopt a sinusoidal features PINN architecture [37] by defining mappings:

$$\gamma(\boldsymbol{v}) = \sin(2\pi(\mathbf{W}\boldsymbol{v} + \mathbf{b})), \tag{3}$$

that act on PINNs' $d$-dimensional spatial-temporal inputs $\boldsymbol{v} = [x,t]^{\mathrm{T}}$. Here, the weights $\mathbf{W}\epsilon\mathbb{R}^{m\times d}$ is a real matrix that maps inputs $\boldsymbol{v}$ into $m$ sinusoidal features and is also related to the frequency of sinusoidal features. The bias $\mathbf{b}\epsilon\mathbb{R}^{m\times 1}$ is a real vector and is also related to phase lag. We incorporate this sinusoidal mapping $\gamma(\boldsymbol{v})$ into the first hidden layer of a PINN and initialize the weights in $\mathbf{W}$ by sampling from the normal distribution $\mathcal{N}(0,\sigma^2)$, $\sigma = 1$. The bias $\mathbf{b}$ is initialized as a zero vector. The subsequent hidden layers also use "sine" activation, and their weights are initialized by He uniform distribution [55,56]. A "linear" activation function is used in the final (output) layer. Moreover, it is recognized that the convergence of stochastic gradient descent methods, including that of ADAM, is highly sensitive to the learning rate. Hence, in present study, a learning strategy to reduce the learning rate on plateauing is adopted to speed up the convergence of ADAM [57].

*2.2. Improve training efficiency with numerical differentiation PINNs (n-PINNs)*

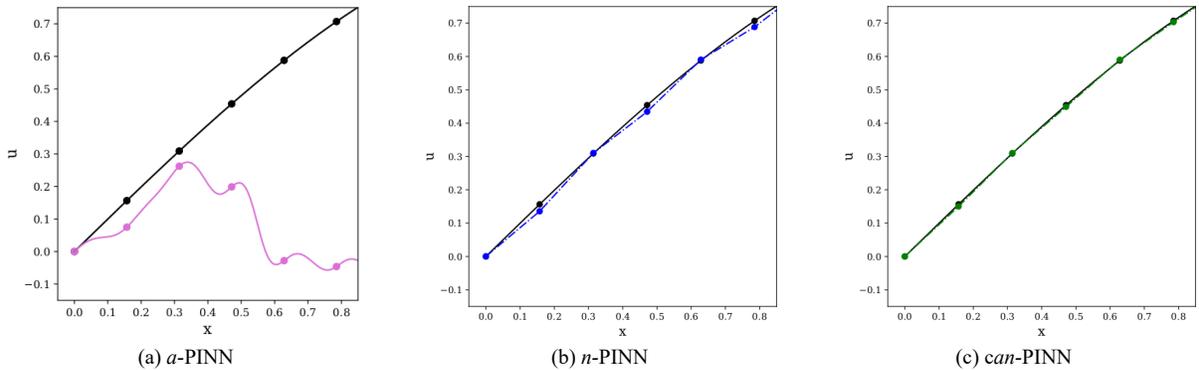

(a) *a*-PINN        (b) *n*-PINN        (c) *can*-PINN

Fig. 1. Schematic diagrams showing (a) *a*-PINN almost perfectly matches the differential operator constraint at all the collocation points (colored) but fails to obtain the true solution (black). Both (b) *n*-PINN and (c) *can*-PINN can approximate the true solution (black) by matching the gradient behavior at the piecewise local regions defined by support points surrounding the collocation points (colored).

It is very common to see a PINN trained in an over-parameterized regime, i.e., by specifying a DNN that has more complexity than the problem requires. If the collocation points sampled during the training are not dense enough, the PINN is susceptible to obtaining an inaccurate or even an obviously unphysical solution. This is particularly true for *a*-PINNs because the AD method computes differential operators exactly at the given collocation point. All collocation points are constrained *almost individually* on a flexible *a*-PINN. The AD-formulated loss function is likely an under-constrained optimization problem when the neural network is heavily over-parameterized. As a result, the *a*-PINN may near perfectly fulfill the underlying differential equation at all the collocation points, leading to a near zero training loss even when its solution is entirely different from the true solution (as illustrated in Figure 1a). The *a*-PINN training therefore becomes completely unrelated to the accuracy of its solution in sparse sample regimes, such that the training loss value can be extremely misleading if one were to apply the *a*-PINNs to a new problem without knowing the ground truth. This is particularly critical as PINN-type methods have been proposed as a mesh-free method to solve complex high-dimensional PDE problems where dense sampling might be impractical [17], and ways to assess the accuracy of the neural network solution besides training loss may not be available. Large amounts of sample points and training iterations—both tend to grow with the problem complexity—and a highly non-trivial task of tuning the training hyper-parameters are required to avoid such under-constrained optimization. The inefficient use of training samples makes *a*-PINN impractical for solving a difficult problem.

To alleviate this issue, we employ ND to replace AD for the computation of differential operators required in PINN training loss. As a very reliable and robust method, numerical differentiation is widely used in scientific computing and computational physics community. The fundamental idea of numerical differentiation is to approximate the derivative terms by means of *local support points*. By choosing proper points and eliminating the leading error terms by utilizing the Taylor-series expansions, the numerical derivative terms can be obtained. A particularly well-known instance of this methodology is the finite-difference method [50].

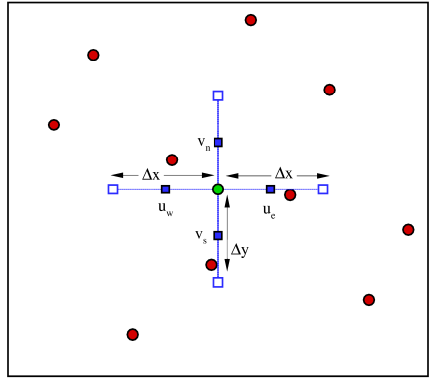

Fig. 2. Schematic diagrams of the definition of the present *n*-PINNs and ca*n*-PINNs framework. The circles represent collocation points, while squares are the additional support points being evaluated by PINN for constructing the derivative terms for a given collocation point (in green).

When approximating the first order derivative $\frac{\partial u(x)}{\partial x}$, the following equation is employed in this study:

$$\frac{\partial u(x)}{\partial x} = \frac{\hat{u}_e - \hat{u}_w}{\Delta x}, \qquad (4)$$

where $\hat{u}_e$ and $\hat{u}_w$ are determined and located at $(x + \frac{\Delta x}{2})$ and $(x - \frac{\Delta x}{2})$. In the above, $\Delta x$ is the distance between two adjacent points for the conventional numerical scheme. Without loss of generality, here we take $\hat{u}_e$ as an example, it may be approximated by 1$^{st}$ order upwind $\left(u_e|_{uw1} = \hat{u}(x; \mathbf{w})\right)$, 2$^{nd}$ order upwind $\left(u_e|_{uw2} = \frac{3}{2}\hat{u}(x; \mathbf{w}) - \frac{1}{2}\hat{u}(x - \Delta x; \mathbf{w})\right)$, or 2$^{nd}$ order central difference $\left(u_e|_{cd2} = \frac{1}{2}\hat{u}(x + \Delta x; \mathbf{w}) - \frac{1}{2}\hat{u}(x; \mathbf{w})\right)$ through Taylor series expansions. For ease of notation, we will reference the use of these conventional numerical schemes for *n*-PINN by *n(uw1)*, *n(uw2)*, and *n(cd2)* in the subsequent sections. The corresponding derivative terms can be derived as:

| Scheme | | Convection term | |
|---|---|---|---|
| 1$^{st}$ order upwind | *n(uw1)* | $\frac{\partial u(x)}{\partial x} \cong \frac{\partial u(x)}{\partial x}\big|_{uw1} = \frac{u_e|_{uw1} - u_w|_{uw1}}{\Delta x} = \frac{\hat{u}(x;\mathbf{w}) - \hat{u}(x-\Delta x;\mathbf{w})}{\Delta x}$, | (5a) |
| 2$^{nd}$ order upwind | *n(uw2)* | $\frac{\partial u(x)}{\partial x} \cong \frac{\partial u(x)}{\partial x}\big|_{uw2} = \frac{u_e|_{uw2} - u_w|_{uw2}}{\Delta x} = \frac{3\hat{u}(x;\mathbf{w}) - 4\hat{u}(x-\Delta x;\mathbf{w}) + \hat{u}(x-2\Delta x;\mathbf{w})}{2\Delta x}$, | (5b) |
| 2$^{nd}$ order central difference | *n(cd2)* | $\frac{\partial u(x)}{\partial x} \cong \frac{\partial u(x)}{\partial x}\big|_{cd2} = \frac{u_e|_{cd2} - u_w|_{cd2}}{\Delta x} = \frac{\hat{u}(x+\Delta x;\mathbf{w}) - \hat{u}(x-\Delta x;\mathbf{w})}{2\Delta x}$. | (5c) |

For *n*-PINNs framework, $\Delta x$ is now a hyper-parameter, and the $\hat{u}$ value at $(x + \Delta x)$ and $(x - \Delta x)$ are obtained by $\hat{u}(x + \Delta x; \boldsymbol{w})$ and $\hat{u}(x - \Delta x; \boldsymbol{w})$, as illustrated in Figure 2. Thus, the present *n*-PINNs framework shares the same appealing feature as *a*-PINNs. They are meshless, i.e., they obtain solutions without the usual mesh generation, and only require a set of collocation points. By virtue of numerical differentiation during the training, *n*-PINNs can consistently obtain reliable solution in both sparse and dense sample regimes. This is because the differential operators computed by numerical differentiation are defined by the *local support points* surrounding the collocation points; in effect, the *n*-PINNs' training aims to modulate the gradient behaviors at piecewise local regions in the solution space, rather than at isolated collocation points. Moreover, while piecewise local regions can be better connected by an appropriate choice of collocation points, it is important to note that the points do not need to be perfectly matching like a typical mesh. As a result, *n*-PINNs can more effectively learn the coherence pattern in the solution even with sparsely sampled collocation points (as illustrated in Figure 1b), leading to a correct solution as the training progresses.

*2.3. Coupled automatic-numerical differentiation PINNs (can-PINNs)*

We then propose to augment the accuracy of *n*-PINNs via the coupled-automatic-numerical differentiation method. Inspired by the multi-moment approach [58–60], the idea of the proposed *can*-PINNs is to approximate the first-order derivative term $\hat{u}_{x|can}$ by virtue of both $\hat{u}$ and $\hat{u}_x$, where $\hat{u}_x$ is obtained from AD, allowing for better gradient-matching behavior in the limit as per Figure 1c.

*2.3.1. Upwind scheme for can-PINNs, can(uw2)*

To couple $\hat{u}$ and $\hat{u}_x$, $\hat{u}_e$ is approximated as $u_e|_{can(uw2)}$ in this study:

$$\hat{u}_e \cong u_e|_{can(uw2)} = a_1 \hat{u}(x; \boldsymbol{w}) + a_2 \hat{u}_x(x; \boldsymbol{w}). \tag{6a}$$

It is followed by performing Taylor series expansions with respect to $\hat{u}_e$ and $\frac{\partial \hat{u}_e}{\partial x}$ for $\hat{u}(x; \boldsymbol{w})$ and $\hat{u}_x(x; \boldsymbol{w})$:

$$\hat{u}(x; \boldsymbol{w}) = \hat{u}_e - \frac{\Delta x}{2}\frac{\partial \hat{u}_e}{\partial x} + \left(\frac{\Delta x}{2}\right)^2 \frac{\partial^2 \hat{u}_e}{\partial x^2} - \left(\frac{\Delta x}{2}\right)^3 \frac{\partial^3 \hat{u}_e}{\partial x^3} + \left(\frac{\Delta x}{2}\right)^4 \frac{\partial^4 \hat{u}_e}{\partial x^4} + \cdots \tag{6b}$$

$$\hat{u}_x(x; \boldsymbol{w}) = \frac{\partial \hat{u}_e}{\partial x} - \frac{\Delta x}{2}\frac{\partial^2 \hat{u}_e}{\partial x^2} + \left(\frac{\Delta x}{2}\right)^2 \frac{\partial^3 \hat{u}_e}{\partial x^3} - \left(\frac{\Delta x}{2}\right)^3 \frac{\partial^4 \hat{u}_e}{\partial x^4} + \left(\frac{\Delta x}{2}\right)^4 \frac{\partial^5 \hat{u}_e}{\partial x^5} + \cdots, \tag{6c}$$

The following two equations can then be derived by eliminating the leading error terms:

$$a_1 = 1 \tag{7a}$$
$$a_2 = \frac{a_1 \Delta x}{2} = \frac{\Delta x}{2}. \tag{7b}$$

The substitution of $a_1$ and $a_2$ in equation (6a) leads to:

$$u_e|_{can(uw2)} = \hat{u}(x; \boldsymbol{w}) + \frac{\Delta x}{2}\hat{u}_x(x; \boldsymbol{w}). \tag{8a}$$

$u_w|_{can}$ can be derived in a similar way as:

$$u_w|_{can(uw2)} = \hat{u}(x - \Delta x; \boldsymbol{w}) + \frac{\Delta x}{2}\hat{u}_x(x - \Delta x; \boldsymbol{w}). \tag{8b}$$

The first order derivative can then be approximated by:

$$\frac{\partial u(x)}{\partial x} \cong \frac{\partial u(x)}{\partial x}\bigg|_{can(uw2)} = \frac{u_e|_{can(uw2)} - u_w|_{can(uw2)}}{\Delta x} = \frac{\hat{u}(x; \boldsymbol{w}) - \hat{u}(x - \Delta x; \boldsymbol{w})}{\Delta x} + \frac{1}{2}\left(\hat{u}_x(x; \boldsymbol{w}) - \hat{u}_x(x - \Delta x; \boldsymbol{w})\right). \tag{9}$$

The modified equation analysis is performed by recasting the above equations as:

$$\frac{\partial u(x)}{\partial x}\bigg|_{can(uw2)} = \hat{u}_x(x; \boldsymbol{w}) + \left(\frac{\hat{u}(x; \boldsymbol{w}) - \hat{u}(x - \Delta x; \boldsymbol{w})}{\Delta x} - \frac{1}{2}\left(\hat{u}_x(x; \boldsymbol{w}) + \hat{u}_x(x - \Delta x; \boldsymbol{w})\right)\right). \tag{10}$$

By performing the Taylor series expansions on $\hat{u}(x - \Delta x; \boldsymbol{w})$ and $\hat{u}_x(x - \Delta x; \boldsymbol{w})$ with respect to $\hat{u}(x; \boldsymbol{w})$ and $\hat{u}_x(x; \boldsymbol{w})$, equation (10) can then be further simplified as:

$$\frac{\partial u(x)}{\partial x}\Big|_{can(uw2)} = \hat{u}_x(x; \boldsymbol{w}) - \left(\frac{\Delta x^2}{12}\right)\hat{u}_{xxx}(x; \boldsymbol{w}) + \frac{\Delta x^3}{24}\hat{u}_{xxxx}(x; \boldsymbol{w}) + \cdots. \qquad (11)$$

From equation (11), the present *can(uw2)* scheme can be seen as including additional stabilization terms to couple the information from adjacent upwind points, and to have a theoretical accuracy order of two. Additionally, in the limit $\Delta x \to 0$, the derivative is exactly equal to the derivative $\hat{u}_x$ as computed by AD.

### 2.3.2. Central scheme for can-PINNs, can(cd)

When solving the incompressible N-S equations with the primitive variables (i.e., velocity and pressure) on collocated points, decoupling of velocity and pressure may happen if one employs the conventional central difference for pressure gradient terms. It is because when approximating the pressure gradient term ($\frac{\partial p(x)}{\partial x}$) by central difference scheme, only adjacent points (($x - \Delta x$) and ($x + \Delta x$)) contribute to the difference formula. To alleviate the decoupling between velocity and pressure, it is crucial to include the contributions at the collocation point $x$, so as to modulate and couple the surrounding pressure field [61,62]. In this study, we propose the following central-based *can*-PINN scheme for approximating the pressure gradient term:

$$\hat{p}_e \cong p_e|_{can(cd)} = \frac{\hat{p}(x + \Delta x; \boldsymbol{w}) + \hat{p}(x; \boldsymbol{w})}{2} - \frac{\Delta x}{8}\left(\hat{p}_x(x + \Delta x; \boldsymbol{w}) - \hat{p}_x(x; \boldsymbol{w})\right) \qquad (12a)$$

$$p_w \cong p_w|_{can(cd)} = \frac{\hat{p}(x; \boldsymbol{w}) + \hat{p}(x - \Delta x; \boldsymbol{w})}{2} - \frac{\Delta x}{8}\left(\hat{p}_x(x; \boldsymbol{w}) - \hat{p}_x(x - \Delta x; \boldsymbol{w})\right), \qquad (12b)$$

and

$$\frac{\partial p(x)}{\partial x} \cong \frac{\partial p(x)}{\partial x}\Big|_{can(cd)} = \frac{p_e|_{can(cd)} - p_w|_{can(cd)}}{\Delta x} = \frac{\hat{p}(x + \Delta x; \boldsymbol{w}) - \hat{p}(x - \Delta x; \boldsymbol{w})}{2\Delta x} - \frac{1}{8}\left(\hat{p}_x(x + \Delta x; \boldsymbol{w}) - 2\hat{p}_x(x; \boldsymbol{w}) + \hat{p}_x(x - \Delta x; \boldsymbol{w})\right). \qquad (13)$$

In the above, it is clearly shown that both adjacent points and collocated contribution $\hat{p}_x(x; \boldsymbol{w})$ are now included in the equation. The above equation (13) can also be recast to show that the proposed *can(cd)* scheme has a theoretical accuracy order of two, by virtue of a modified equation analysis as described in section 2.3.1:

$$\frac{\partial p(x)}{\partial x}\Big|_{can(cd)} = \hat{p}_x(x; \boldsymbol{w}) + \left(\frac{\hat{p}(x + \Delta x; \boldsymbol{w}) - \hat{p}(x - \Delta x; \boldsymbol{w})}{2\Delta x} - \frac{1}{8}\left(\hat{p}_x(x + \Delta x; \boldsymbol{w}) + 6\hat{p}_x(x; \boldsymbol{w}) + \hat{p}_x(x - \Delta x; \boldsymbol{w})\right)\right), \qquad (14a)$$

$$\frac{\partial p(x)}{\partial x}\Big|_{can(cd)} = \hat{p}_x(x; \boldsymbol{w}) + \left(\frac{\Delta x^2}{24}\right)\hat{p}_{xxx}(x; \boldsymbol{w}) - \frac{\Delta x^4}{480}\hat{p}_{xxxxx}(x; \boldsymbol{w}) + \cdots. \qquad (14b)$$

Note that although this work focuses on the above common schemes, the *n-/can*-PINN framework can be extended to other schemes which are based on Taylor series expansion of varying form and accuracy.

### 2.4. Fundamental analysis

This subsection investigates the dispersion and dissipation behavior for the proposed *can(uw2)* and *can(cd)* schemes. The analysis starts by assuming that Fourier transform and its inverse can be employed to a field variable $\phi$:

$$\tilde{\phi}(\alpha) = \frac{1}{2\pi}\int_{-\infty}^{\infty} \phi(x)\exp(-i\alpha x)\,dx \qquad (15a)$$

$$\phi(x) = \frac{1}{2\pi}\int_{-\infty}^{\infty} \tilde{\phi}(\alpha)\exp(i\alpha x)\,d\alpha. \qquad (15b)$$

It is followed by employing the Fourier transform to the following difference equation $\frac{\partial \phi}{\partial x}\Big|_{can(uw2)}$ with $\Delta x = h$:

$$\frac{\partial \phi(x)}{\partial x} \cong \frac{\partial \phi(x)}{\partial x}\Big|_{can(uw2)} = \frac{\phi(x) - \phi(x - h)}{h} + \frac{1}{2}\left(\phi_x(x) - \phi_x(x - h)\right). \qquad (16)$$

In the above, $\phi_x$ may be obtained by AD. It will lead to

$$i\alpha h \cong (1 - \exp(-i\alpha h)) + \frac{i\alpha h}{2}(1 - \exp(-i\alpha h)). \tag{17}$$

By defining effective wavenumber $\alpha' \cong \alpha$, the above equation can be re-derived as

$$\alpha' h = -i\left((1 - \exp(-i\alpha h)) + \frac{i\alpha h}{2}(1 - \exp(-i\alpha h))\right). \tag{18}$$

It can also be done for the difference equation $\phi_{x|can(cd)}$ in a similar way:

$$\frac{\partial \phi(x)}{\partial x} \cong \frac{\partial \phi(x)}{\partial x}\Big|_{can(cd)} = \frac{\phi(x+h)+\phi(x-h)}{2h} - \frac{1}{8}\left(\phi_x(x+h) - 2\phi_x(x) + \phi_x(x-h)\right), \tag{19a}$$

$$i\alpha h \cong \frac{(\exp(i\alpha h) - \exp(-i\alpha h))}{2} - \frac{i\alpha h}{8}(\exp(i\alpha h) - 2 - \exp(-i\alpha h)), \tag{19b}$$

$$\alpha' h = -i\left(\frac{(\exp(i\alpha h) - \exp(-i\alpha h))}{2} - \frac{i\alpha h}{8}(\exp(i\alpha h) - 2 - \exp(-i\alpha h))\right). \tag{19c}$$

With the definition of dispersion ($k_i$) and dissipation ($k_r$),

$$k_i = \Re(\alpha' h) \tag{20a}$$
$$k_r = \Im(\alpha' h), \tag{20b}$$

where $\Re$ and $\Im$ are the real and imaginary components, we plot the $(\alpha h - k_i)$ and $k_r$ against modifed wavenumber $\alpha h$ in Figure 3 to visually show the dispersion and dissipation behavior for the newly proposed schemes. It can be seen from Figure 3(a) that both the proposed *can(uw2)* and *can(cd)* have smaller dispersion errors than conventional 1st and 2nd order upwind schemes. Figure 3(b) also shows that *can(uw2)* has smaller dissipation, which increases smoothly for higher $\alpha h$, while *can(cd)* is a non-dissipative scheme. Hence, it is expected that the proposed *can(uw2)* and *can(cd)* schemes can perform better for more dispersive problems in comparison to the baseline numerical schemes.

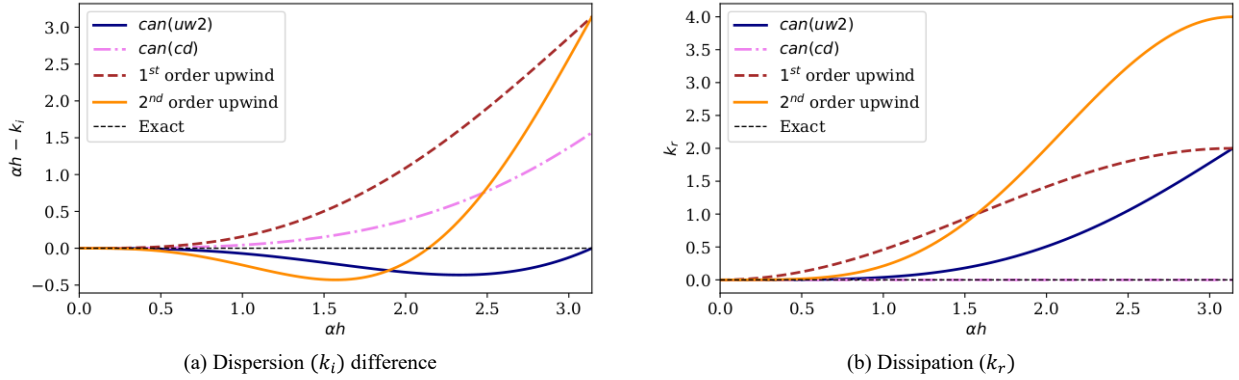

(a) Dispersion ($k_i$) difference      (b) Dissipation ($k_r$)

Fig. 3. Plots of fundamental analysis for the proposed *can(uw2)* and *can(cd)*, and the conventional 1st and 2nd order upwind schemes.

## 3. Experimental study

The general PINN architecture and training configurations used in our experimental study are summarized in Table 1. For each test problem, we compare the performance of *a*-PINNs, *n*-PINNs, and *can*-PINNs, using identical network architecture and training setting. We also report their training cost.

Table 1. PINN architecture and training configurations used in the experimental study and their training cost.

| Problem | 3.1. ODE | 3.2. Flow mixing | 3.3. Lid-driven cavity | 3.4. Backward facing step | 3.5. Complex channel |
|---|---|---|---|---|---|
| Governing eqns. | (21) | (22) | (27) | (27) | (27) |
| PINN architecture | $(x)$–64–20–20–20–$(\hat{u})$ | $(x,y,t)$–64–20–20–20–$(\hat{u})$ | $(x,y)$–64–20–20–20–[20–20–20–$(\hat{u})$, 20–20–20–$(\hat{v})$, 20–20–20–$(\hat{p})$] | $(x,y)$–64–20–20–20–[20–20–20–$(\hat{u})$, 20–20–20–$(\hat{v})$, 20–20–20–$(\hat{p})$] | $(x,y)$–128–30–30–30–[30–30–30–$(\hat{u})$, 30–30–30–$(\hat{v})$, 30–30–30–$(\hat{p})$] |
| $\Delta x$ ($n$- & $can$-PINNs) | $\frac{2\pi}{40}$ / $\frac{2\pi}{80}$ / $\frac{2\pi}{60}$ | $\Delta x = \Delta y = 0.16$ | $\Delta x = \Delta y = 0.02$ | $\Delta x = 0.05, \Delta y = 0.026$ | $\Delta x = \Delta y = 0.02$ |
| Training sample (total number of collocation points) | 41 / 81 / 161 | 65,025 | 2,601 | 16,000 | 3,437 |
| Batch size | 6 + 0 + 2 | 470 + 15 + 15 | 475 + 0 + 25 | 475 + 0 + 25 | 475 + 0 + 25 |
| Max. training iteration | 100,000 | 100,000 | 200,000 | 500,000 | 500,000 |
| Initial learning rate | 5e-3 | 5e-3 | 1e-3 | 1e-3 | 1e-3 |
| Total training cost ($a$-/$n$-/$can$-PINNs) | 5.1 / 5.0 / 5.5 mins | 10.9 / 8.4 / 9.3 mins | 28.3 / 24.2 / 29.2 mins | 72.6 / 63.4 / 74.7 mins | 82.9 / 54.1 / 77.3 mins |

1. For the PINN architecture, the numbers in between input and output represent the number of nodes in each hidden layers. For example, $(x)$–64–20–20–20–$(\hat{u})$ indicates a single input $x$, followed by 4 hidden layers with 64, 20, 20 and 20 nodes in each layer, and a single output $\hat{u}$.
2. We incorporate the sinusoidal mapping (3) into the first hidden layer of PINN and initialize its weights by sampling from a normal distribution $\mathcal{N}(0, \sigma^2)$, $\sigma = 1$. The subsequent hidden layers use "sine" activation, except a "linear" activation function is used in the final (output) layer, and their weights are initialized by He uniform distribution.
3. Batch size: number of collocation points sampled for 1 evaluation of $\mathcal{L}_{PINN} = \lambda^{-1}\mathcal{L}_{PDE} + \mathcal{L}_{IC} + \mathcal{L}_{BC}$. We used a default $\lambda=1$.
4. A training iteration: 1 evaluation of $\mathcal{L}_{PINN}$ for backpropagating the weight gradients. We update PINN weights in every 100 iterations. We reduce the learning rate on plateauing, until a min. learning rate of 5e-6 is reached.
5. Training cost are compared on a 20-core workstation with the Intel Xeon Gold 6248 processors.

### 3.1. Validation on ODE

In this section, the following ODE

$$\frac{du}{dx} = f, \qquad (21)$$

will be solved by the proposed *can*-PINN (*can(uw2)* and *can(cd)*) framework in a 1D domain of $x = (0, 2\pi)$. The ODE is investigated with two different source terms: $f_1 = \cos(x)$ and $f_2 = \cos(x) + 2\cos(2x)$. With the boundary condition $u(0) = u(2\pi) = 0$, the corresponding ground truth solutions for the investigated ODE are $u = \sin(x)$ and $u = \sin(x) + \sin(2x)$, respectively. The training batch size is set as 6 and total training iterations are 1e5 to ensure the PINN training is fully converged. We solve the two ODE problems with 41, 81, and 161 equidistantly spaced collocation points, and compare the performance of *a*-PINN with the proposed *n*-PINN and *can*-PINN utilizing 1st order upwind, *n(uw1)*, 2nd order upwind, *n(uw2)* and *can(uw2)*, and central difference, *can(cd)*, schemes. For each model, we perform 10 independent runs and plot the distribution of mean square error (MSE) between their solutions and the ground truth.

#### 3.1.1. Comparison between a-PINN, n-PINN, and can-PINNs results

Figure 4 shows that *a*-PINN solutions to the two ODE problems with 41 collocation points have significantly large MSE, even though the training loss decreases to an extremely low value, i.e., below 1e-7. Both *n*-PINNs and *can*-PINNs can approximate the true solutions with an accuracy that is between 1 to 3 orders better. On the other hand, when the collocation points increase to 81 and even 161, *a*-PINN can produce more accurate solutions and outperform *n*-PINNs. All three PINNs can produce more accurate solutions with more samples, and the solutions obtained by *can*-PINNs are statistically better than *n*-PINNs for all scenarios. It is also noticed that even though *can*-PINNs are theoretically second order accurate, with the inclusion of AD into the scheme, we can get more accurate solutions than 2nd order upwind scheme, *n*(uw2). It is believed that this is because the proposed schemes can achieve better dispersion behavior, as well as have smaller leading error coefficients. The solution at 50[th] percentile MSE for each of the PINN models obtained when solving for problem $\frac{du}{dx} = \cos(x)$ with 41 collocation points are plotted in Figure 5. Compared with the ground truth, the *a*-PINN fails to solve the problem. The solution from *n(uw1)*-PINN also has obvious discrepancy. The solutions from *n(uw2)*-PINN and *can(cd)*-PINN are very close to the ground truth, while that from *can(uw2)*-PINN almost overlaps with the ground truth.

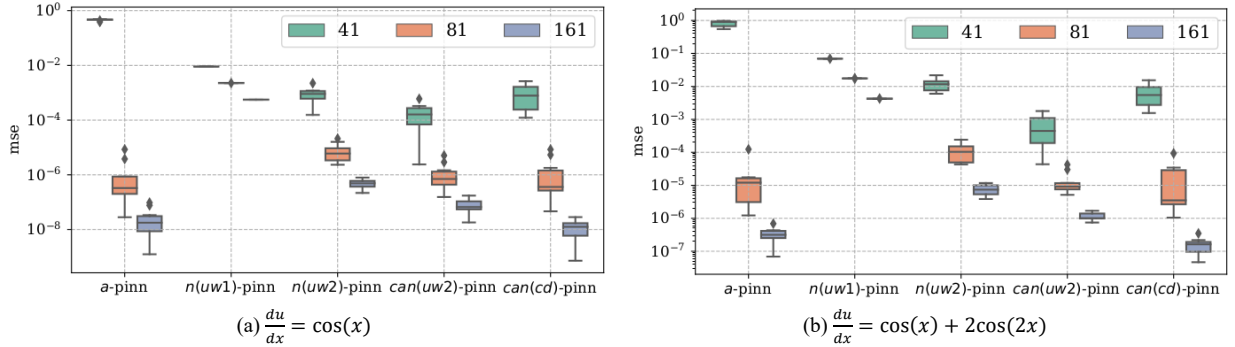

(a) $\frac{du}{dx} = \cos(x)$  (b) $\frac{du}{dx} = \cos(x) + 2\cos(2x)$

Fig. 4. Distribution of MSE between PINNs (*a*-PINNs, *n*-PINNs, and *can*-PINNs) and ground truth solutions for the ODE problems. Results from 10 independent runs are shown as boxplot. All the models have a training loss below 5e-6.

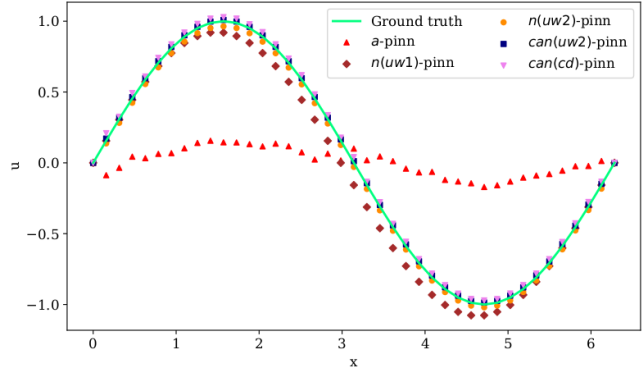

Fig. 5. Comparison between the ground truth solution and the solutions solved by *a*-PINN, *n*-PINNs, and *can*-PINNs, for the ODE problem $\frac{du}{dx} = \cos(x)$ with 41 collocation points. For each PINN model, the solution at 50[th] percentile MSE from 10 independent runs is shown.

## 3.2. Flow mixing

To further illustrate the advantages of the *n*-PINN and *can*-PINN frameworks, we then move on to the two-dimensional flow-mixing problem. In the spatial domain $x \in [-4, 4], y \in [-4, 4]$, two fluids with different properties are mixed at the interface by a specified rotational velocity $v_t$ [63]. The governing equation for this transient problem is written as

$$\frac{\partial u}{\partial t} + a \frac{\partial u}{\partial x} + b \frac{\partial u}{\partial y} = 0, \tag{22}$$

where

$$a(x, y) = -\frac{v_t}{v_{tmax}} \frac{y}{r} \tag{23a}$$
$$b(x, y) = \frac{v_t}{v_{tmax}} \frac{x}{r}, \tag{23b}$$
$$v_t = \text{sech}^2(r) \tanh(r) \tag{23c}$$
$$r = \sqrt{x^2 + y^2}, \tag{23d}$$

The corresponding analytical solution is:

$$u(x, y, t) = -\tanh\left(\frac{y}{2}\cos(\omega t) - \frac{x}{2}\sin(\omega t)\right), \quad (24)$$

where $\omega = \frac{1}{r}\frac{v_t}{v_{tmax}}$ [63]. It is noted that in this study, $v_{tmax}$ is set as 0.385. We solve the problem with PINNs for $t \in [0, 4]$, with the initial condition at $t = 0$ and Dirichlet boundary condition at the spatial boundaries, as specified by equation (24).

To employ the *n*-PINN and *can*-PINNs, the governing equation (22) is recast as the conservative form:

$$\frac{\partial u}{\partial t} + \frac{\partial (au)}{\partial x} + \frac{\partial (bu)}{\partial y} = u\left(\frac{\partial a}{\partial x} + \frac{\partial b}{\partial y}\right). \quad (25)$$

The spatial derivative terms can be derived based on the approach described in Section 2.3

$$\frac{\partial (au)}{\partial x} = \frac{a_e u_e - a_w u_w}{\Delta x} \quad (26a)$$

$$\frac{\partial (bu)}{\partial y} = \frac{b_n u_n - b_s u_s}{\Delta y}, \quad (26b)$$

where $u_e, u_w, u_n$ and $u_s$ are approximated by the 2$^{nd}$ order upwind scheme *n*-PINN, *can(uw2)*- and *can(cd)*-PINNs. The temporal derivative term in (25) is obtained by AD.

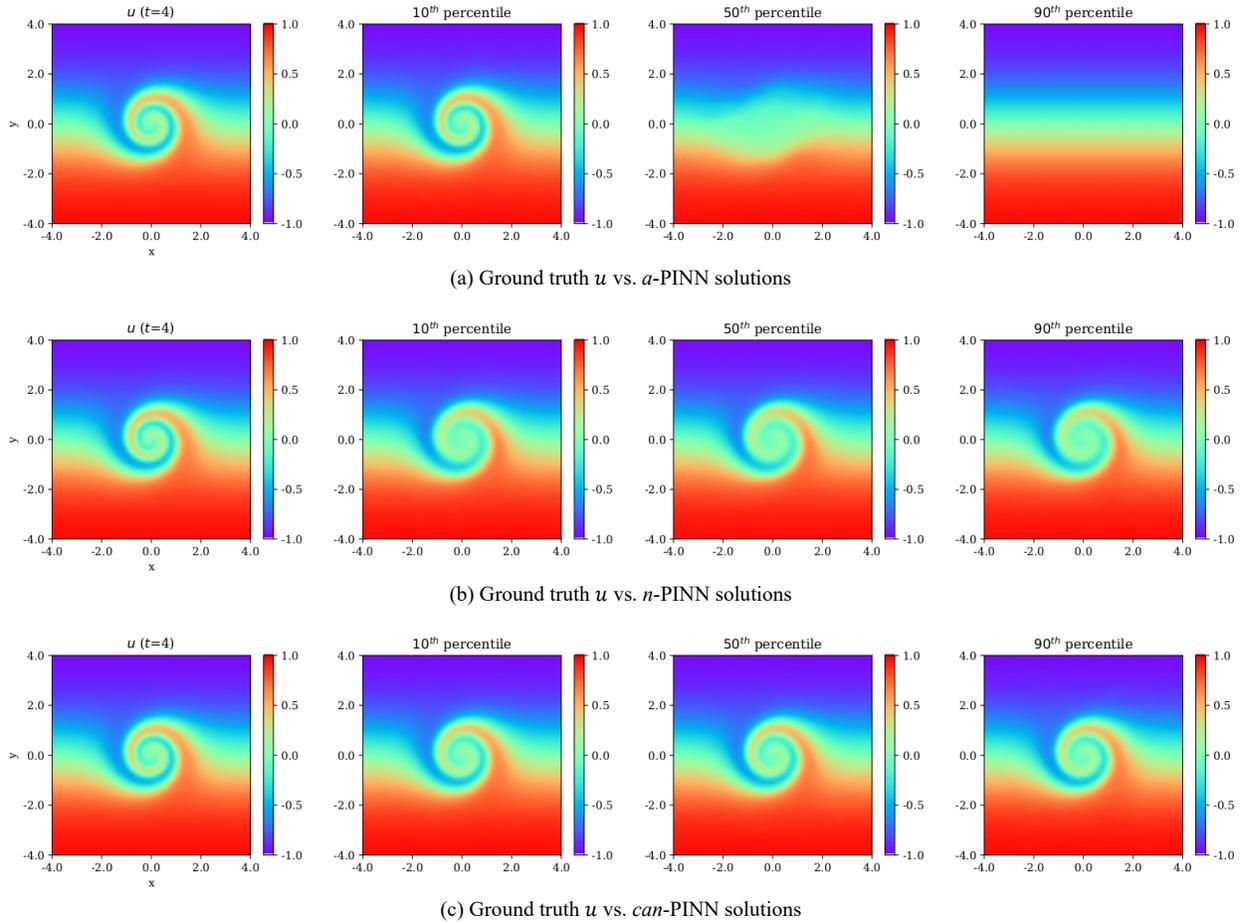

Fig. 6. Comparison between the ground truth solution ($t = 4$) and the solutions solved by (a) *a*-PINN, (b) *n*-PINN, and (c) *can(uw2)*-PINN, for the flow mixing problem. The solutions on 10$^{th}$, 50$^{th}$, and 90$^{th}$ percentiles of ascending MSE from 50 independent runs are shown.

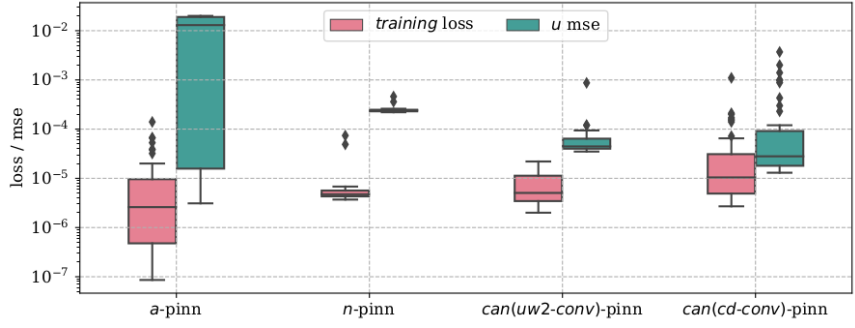

Fig. 7. Distribution of training loss and MSE between PINNs (*a*-PINNs, *n*-PINNs, and *can*-PINNs) and ground truth solutions for the flow mixing problem. Results from 50 independent runs are shown as boxplot.

*3.2.1. Comparison between a-PINN, n-PINN, and can-PINNs results*

We compare the performance of *a*-PINN, *n*-PINN, and *can*-PINNs (both *can(uw2)* and *can(cd)* schemes), by training the models with a set of 65,025 collocation points. The PINN architecture and training settings are identical for all models, as listed in Table 1. For each model, we perform 50 independent runs. Figure 6 visually compares the 10$^{th}$, 50$^{th}$, and 90$^{th}$ percentile solutions obtained by *a*-PINN, *n*-PINN, and *can(uw2)*-PINN, of ascending MSE, while the distribution of training losses and their solutions' MSE against the ground truth are shown in Figure 7. The results clearly show that *a*-PINN is unable to consistently produce accurate solution, causing a very large spread in its MSE distribution, even though their training losses tend to be the lowest among the different PINN schemes. Hence the low training loss given by *a*-PINN can be very misleading. On the other hand, both *n*-PINN and *can*-PINN are able to consistently obtain the correct flow pattern, as demonstrated in Figure 6. Comparing their training loss and MSE distributions in Figure 7, the solutions produced by both *can(uw2)*-PINN and *can(cd)*-PINN are significantly more accurate than *n*-PINN. It is also noticed that the *can(cd)*-PINN has the ability to obtain more accurate solution than *can(uw2)*-PINN. However, due to non-dissipative nature of central scheme, the *can(cd)*-PINN has larger variance than the latter scheme. Overall, the upwind-based *can(uw2)*-PINN may be more favorable for this problem, due to its consistency. The efficacy and accuracy of the proposed coupled schemes are thus demonstrated in this test problem.

*3.3. Lid-driven cavity*

The lid-driven cavity problem has been widely chosen as a benchmark case for many numerical methods, due to the complex physics encapsulated within. As per the schematic in Figure 8, this problem is a unit square cavity with a lid velocity $u_{lid} = 1$ for the top wall, while other walls are non-slip. When the Reynolds number ($Re$) is less than 1000, there will only be two eddies at the bottom-right and bottom-left regions. With increasing values of $Re$ to 2500, additional eddies can be observed at top-left regions [64]. When $Re$ is even higher, more eddies will appear.

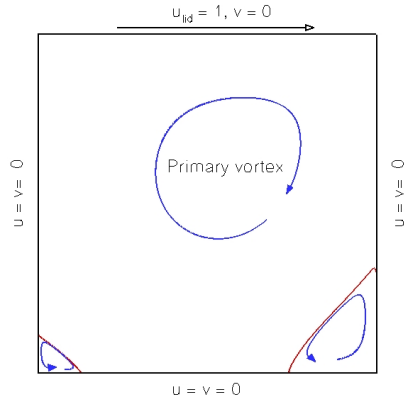

Fig. 8. Schematic of lid-driven cavity problem ($Re = 400$).

The governing equations for this problem are the steady-state, two-dimensional incompressible N-S equations:

$$\frac{\partial u}{\partial x} + \frac{\partial v}{\partial y} = 0 \tag{27a}$$

$$\frac{\partial (uu)}{\partial x} + \frac{\partial (vu)}{\partial y} = \frac{1}{Re}\left(\frac{\partial}{\partial x}\left(\frac{\partial u}{\partial x}\right) + \frac{\partial}{\partial y}\left(\frac{\partial u}{\partial y}\right)\right) - \frac{\partial p}{\partial x} \tag{27b}$$

$$\frac{\partial (uv)}{\partial x} + \frac{\partial (vv)}{\partial y} = \frac{1}{Re}\left(\frac{\partial}{\partial x}\left(\frac{\partial v}{\partial x}\right) + \frac{\partial}{\partial y}\left(\frac{\partial v}{\partial y}\right)\right) - \frac{\partial p}{\partial y}. \tag{27c}$$

In the above equations, the primitive variables $(u, v)$ and $p$ are velocity and pressure, $Re$ is Reynolds number which represents the ratio of inertial forces to viscous forces. To better approximate different derivative terms by taking their physical nature into account, convection terms are approximated by the upwind-based *can(uw2)*, while *can(cd)* scheme is employed for pressure gradient terms for the present *can*-PINNs. Similarly, for *n*-PINNs, a second order upwind scheme and a central difference scheme are employed for convection terms and pressure gradient terms respectively. All other differential terms are approximated by a central difference scheme in both *can*-PINNs and *n*-PINN. In order to better understand the efficacy of employing *can(cd)* for pressure gradient terms (referred to as *can(uw2-conv, cd-p)*-PINN), we also train a *can*-PINN which uses a central difference numerical scheme for pressure gradient terms (referred to as *can(uw2-conv)*-PINN).

We solve for the lid driven cavity problem at $Re = 400$, with a 1×1 unit spatial domain as specified by $x \in [0,1], y \in [-1, 0]$. The PINN architecture and training setting are identical for all models, as listed in Table 1. To compute the MSE for the solution obtained by PINNs, the ground truth is obtained by an in-house numerical solver based on the improved divergence-free condition compensated (IDFC) method [65]. It has been shown in [65] that the IDFC method is reliable and accurate.

*3.3.1. Comparison between a-PINN, n-PINN, and can-PINN results*

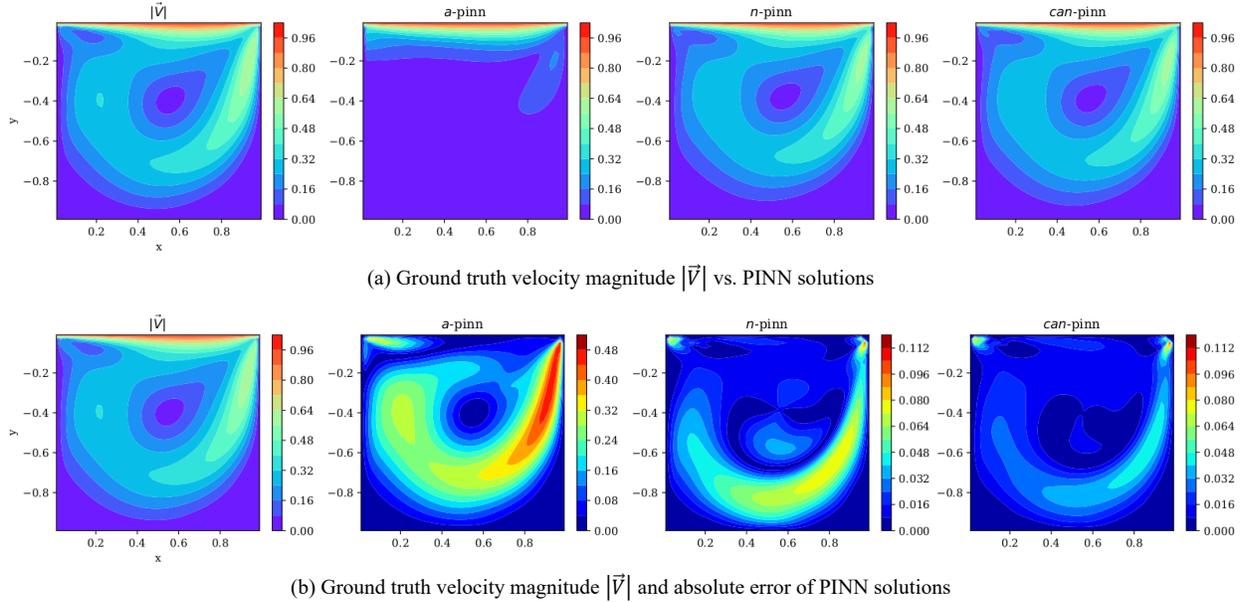

Fig. 9. (a) Comparison between the ground truth velocity magnitude $|\vec{V}|$ and the solutions solved by *a*-PINN, *n*-PINN, and *can(uw2-conv, cd-p)*-PINN, for the lid driven cavity problem at $Re = 400$. (b) The absolute error between the PINN solutions and ground truth. From 50 independent runs, the solution with median MSE for respective PINN models are shown.

We demonstrate the performance of *a*-PINN, *n*-PINN, and *can*-PINNs when trained with 2,601 equidistantly spaced collocation points. For each PINN model, we perform 50 independent runs. Figure 9 visually compares the velocity magnitude $|\vec{V}| = \sqrt{u^2 + v^2}$ contour computed from solutions solved by *a*-PINN, *n*-PINN, and *can(uw2-conv, cd-p)*-PINN, and also their absolute deviation from the simulated ground truth solution. Clearly, it is difficult to obtain a reasonable solution by *a*-PINN with the current 2,601 collocation points, i.e., the correct flow does not develop. While both the *n*-PINN and *can*-PINN show good agreement with the ground truth in their solutions, the *can*-PINN is more accurate. Figure 10a compare their distributions of training loss and solution MSE. Despite having the lowest training loss, the *a*-PINN's solutions are consistently bad. Their MSEs (>1e-2) are more than 1 order of magnitude higher than those obtained by *n*-PINN and is about 2 orders of magnitude higher than *can*-PINNs. The results also show that the proposed *can*-PINNs are significantly more accurate than *n*-PINN. It is also noticed that by utilizing *can(cd)* scheme for the pressure gradient term, the *can(uw2-conv, cd-p)*-PINN's solution has a slightly lower minimum and median MSE, as compared to *can(uw2-conv)*-PINN, further illustrating the advantage of using *can* scheme for pressure gradient.

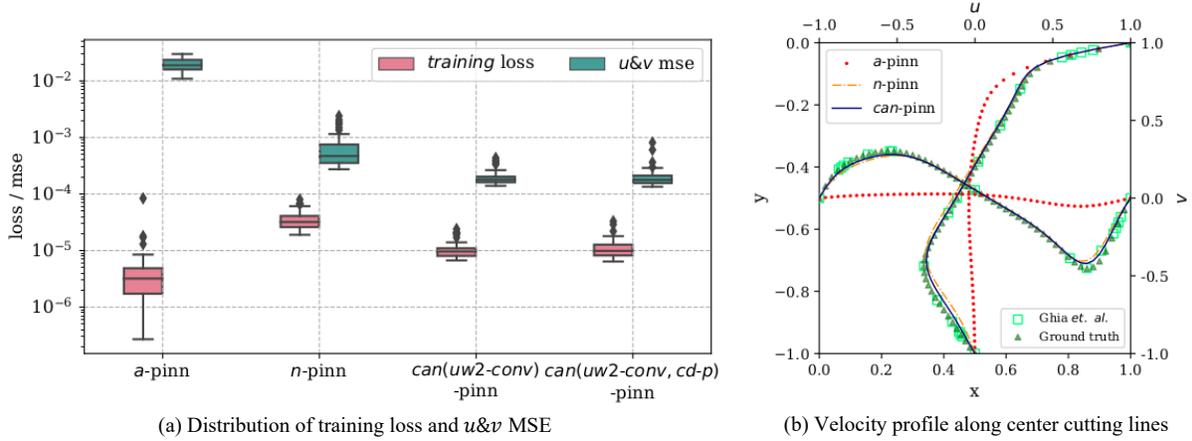

(a) Distribution of training loss and $u\&v$ MSE    (b) Velocity profile along center cutting lines

Fig. 10. (a) Distribution of training loss and $u\&v$ MSE between PINNs (*a*-PINN, *n*-PINN, and *can*-PINNs) and ground truth solutions for the lid driven cavity problem at $Re = 400$. For $u\&v$ MSE, we compute the MSE for $u$- and $v$-velocity components and take the average. Results from 50 independent runs are shown as boxplot. (b) The velocity profiles $u(0.5, y)$ and $v(x, -0.5)$ along the center cutting lines, from the median solution obtained by *a*-PINN, *n*-PINN, and *can(uw2-conv, cd-p)*-PINN vs. the ground truth (simulation) and benchmark results from Ghia *et. al.* [66].

In Figure 10b, we plot the velocity profiles ($u(0.5, y)$ and $v(x, -0.5)$) along the cutting lines at the center of the cavity $x = 0.5$ and $= -0.5$, based on the median solution obtained by *a*-PINN, *n*-PINN, and *can(uw2-conv, cd-p)*-PINN. Good agreement is revealed for the proposed *can*-PINN result with both our in-house simulation result and benchmark result from Ghia *et. al.* [66]. The *n*-PINN velocity profiles deviate slightly more from the simulation and benchmark results, while *a*-PINN results display a large discrepancy.

*3.3.2. Training a-PINN, n-PINN, and can-PINN under different sampling scenarios*

Table 2. The PINN training setting under different sampling scenarios.

| Sampling scenario (collocation points sampled from) | **2,601 equidistantly spaced points** | **10,201 equidistantly spaced points** | **40,401 equidistantly spaced points** | **Uniform distribution** |
|---|---|---|---|---|
| PINN training setting | 200,000 iterations 500 mini-batch pts. | 1,000,000 iterations 500 mini-batch pts. | 2,000,000 iterations 1000 mini-batch pts. | 2,000,000 iterations 1000 mini-batch pts. |
| $\Delta x$ (for *n*- & *can*-PINNs) | $\Delta x = \Delta y = 0.02$ | $\Delta x = \Delta y = 0.01$ | $\Delta x = \Delta y = 0.005$ | $\Delta x = \Delta y = 0.01$ |

As the results in previous sub-section suggest that sampling from a set of 2,601 collocation points is insufficient for training a good *a*-PINN model, we further study the performance of different PINNs under different sampling scenarios for the same test problem. In particular, we train *a*-PINN, *n*-PINN, and *can(uw2-conv, cd-p)*-PINN models with a larger set of equidistantly spaced collocation points, i.e., 10,201 and 40,401. We also adjust the training iteration and batch size, as shown in Table 2, to ensure the training is converged across different sampling scenarios. The distribution of solution MSE based on 10 independent runs for different PINN models and sampling scenarios are presented in Figure 11(a). The results show that all 3 PINN models can achieve a more accurate solution (i.e., more than 1 order of magnitude lower in MSE), when trained with the largest set of 40,401 collocation points, albeit with a tradeoff of more training iteration and batch size. The *a*-PINN is still unable to obtain a reasonable solution with 10,201 collocation points, and finally achieves a good solution with MSE (~2e-4) with 40,401 collocation points. Even then, this result by *a*-PINN is only on par with the results obtained by *can*-PINN with 2,601 collocation points. In addition, this requires both double the mini-batch size and much more training iterations than the sparsely sampled *can*-PINN. Under the same training setting with 40,401 collocation points as the *a*-PINN, the quality of *can*-PINN solutions further improve to an MSE below 1e-5.

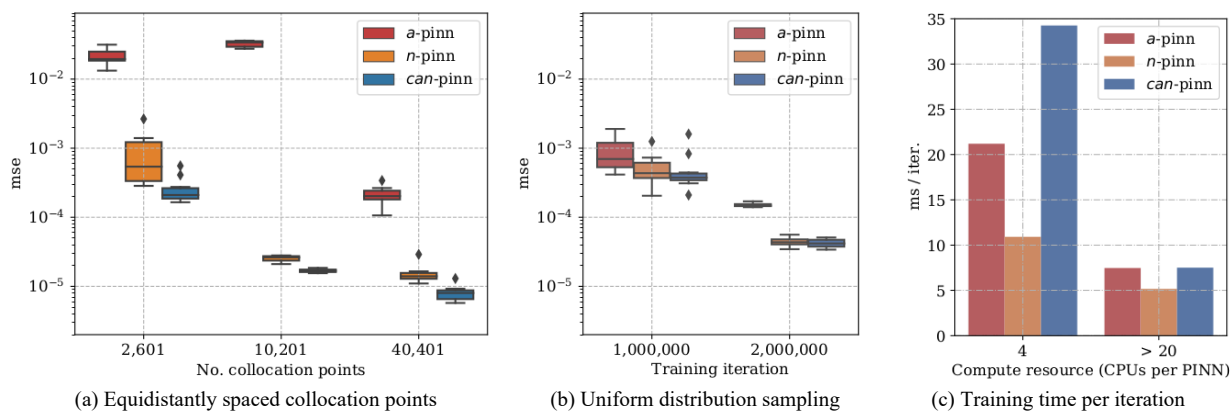

Fig. 11. (a) Distribution of $u\&v$ MSE between PINNs (*a*-PINN, *n*-PINN, and *can(uw2-conv, cd-p)*-PINN) and ground truth solutions for the lid driven cavity problem at $Re = 400$, when trained with 2,601, 10,201 and 40,401 equidistantly spaced collocation points. (b) Distribution of $u\&v$ MSE vs. training iteration, with the collocation points randomly sampled from uniform distribution. For $u\&v$ MSE, we compute the MSE for $u$- and $v$-velocity components and take the average. All the results are aggregated from 10 independent runs. (c) Comparison of PINNs' training time (averaged from 5,000 iterations) under limited and excessive compute resource scenarios. In each iteration, PINN loss is evaluated on 500 mini-batch samples randomly drawn from 2,601 collocation points.

The performance of different PINN models when trained on collocation points randomly sampled from a uniform distribution is also studied (Table 2). For this work, we found $\Delta x = 0.01$ yields a satisfactory performance, although further optimization may generally be necessary for other problems. In the limit as training iteration grows, this random uniform sampling is equivalent to an infinitely dense set of collocation points. Figure 11(b) compares the distribution of solution MSE between *a*-PINN, *n*-PINN, and *can(uw2-conv, cd-p)*-PINN models, at 1 and 2 million training iterations. Again, all 3 PINN models can achieve a more accurate solution after more training iterations. Comparing Figure 11(a) and Figure 11(b), it is noticed that *a*-PINN performs better on the uniform distribution sampling as compared to training with a sparser set of fixed collocation points, given the same large amount of training iteration. On the contrary, *n*-PINN and *can*-PINN training is consistently more efficient under the fixed collocation points sampling scenario for the present test problem. Critically, the proposed *can*-PINN consistently outperforms *a*-PINN and *n*-PINN across all sampling scenario evaluated.

The above results indicate that PINN models can generally achieve a better solution with increased sampling resolution in either uniform or equidistantly spaced sampling scenarios, however, this typically requires longer training. The training efficiency is further impeded by the highly non-trivial task of optimizing the training hyper-parameters, which is usually a practical bottleneck, especially as required training iterations increase and variance in the optimization outcomes increase. With the ability to efficiently train with a sparse set of collocation points while robustly producing an accurate solution, *can*-PINN can potentially solve more challenging PINN problems where previously infeasible with the typical *a*-PINN.

In addition, we report the per iteration training time from different PINN models as used in the present study as per Figure 11(c) for reference. The PINN implementations used the Keras API as packaged with TensorFlow2.5 [67]. During training, *n*-PINN requires only forward pass for the computation of the differential operators for loss evaluation, which can be faster than the back-propagation AD computation utilized by *a*-PINN. The *can*-PINN however performs both forward pass and back-propagation during the loss evaluation. Hence, with a limited compute resource (i.e., 4 CPUs per PINN), the *n*-PINN is the quickest, while the *can*-PINN is the slowest. In addition, Keras and its backend TensorFlow automatically parallelize execution. Hence, when there are over 20 CPUs per PINN, the *a*-PINN, *n*-PINN, and *can*-PINN show similar execution times per iteration. Critically, we note that the *n*-PINNs and *can*-PINNs are also generally more sample efficient and converge with less total iterations than the *a*-PINNs, in addition to having similar execution times per iteration.

*3.3.3. Solving inverse problem with can-PINN*

We further demonstrate the capability of the proposed *can*-PINNs on inverse problem. In an inverse problem, there are certain unknowns in the formulation of the differential equation or initial and boundary conditions, but the outcome of differential equation is partially available in the form of observation data. In particular, we seek to infer the unknown $Re$ in the incompressible N-S equations (27), as well as the solution over the problem domain, by training a *can(uw2-conv, cd-p)*-PINN model with respect to the data-constrained loss function (2). In addition to the known boundary condition as indicated in Figure 8, we assume the availability of *very limited* observations ($n=10$) of velocity (i.e., $u$ and $v$ values obtained from the IDFC simulations). The experiment comprises of 20 independent runs, where observations are randomly drawn. Selected observation sets are displayed in Figure 12a. Besides these observations, the PINN models are trained with 2,601 equidistantly spaced collocation points, following the same training setting listed in Table 1. However, to ensure sufficiently low data loss, we reweight the PINN loss terms to give more priority to the data loss (i.e., the weight of data loss to PDE loss is 100 to 1).

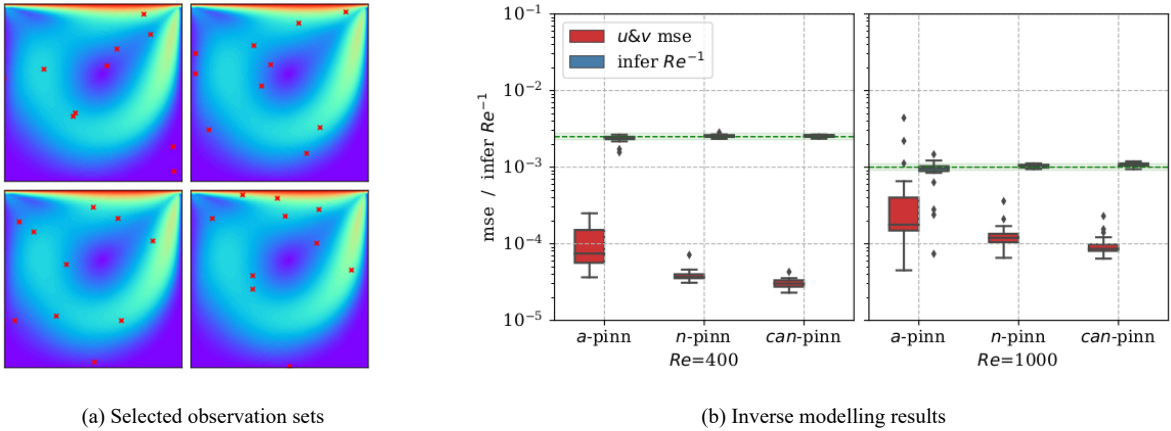

(a) Selected observation sets  (b) Inverse modelling results

Fig. 12. (a) 4 random observation sets in the inverse lid driven cavity problem, where the *can(uw2-conv, cd-p)*-PINN is applied to simultaneously infer the unknown $Re$ coefficient in the incompressible N-S equations and also to solve the solution based on limited observations ($n=10$, in red). (b) The distribution of $u\&v$ MSE and inferred $Re^{-1}$ obtained from the *can(uw2-conv, cd-p)*-PINN for inverse LDC problem at $Re=400, 1000$. The green dashed lines and shaded areas indicate the ground truth $Re^{-1}$ and their 10% error bounds. Results are based on 20 independent runs.

The inverse modelling results are visualized in Figure 12b, comprising the distribution of $u\&v$ MSE and inferred $Re^{-1}$ for both $Re = 400$ and $Re = 1000$ cases. It is observed that the *can*-PINN can accurately infer the unknown $Re$ for both cases. For example, in the $Re = 400$ case, our inverse *can*-PINN model consistently infers an accurate $Re$ value which is always within 10% error from the ground truth and achieve solution MSE below 5e-5. The efficacy drops slightly for the $Re = 1000$ case due to the more complex fluid phenomenon, however, all inferred $Re$ values are still within 20% of the ground truth, and the solutions still achieve a MSE below 5e-4. This further demonstrates the effectiveness of *can*-PINN for solving complex inverse problems from very limited random observations ($n=10$), with consistent performance across different sets of random observation data. Figure 12b also compares the performance of *can*-PINN with *a*-PINN and *n*-PINN. We observed a similar performance from *n*-PINN and *a*-PINN in this inverse problem, although *a*-PINN has the biggest spread in inferred $Re$ and solution MSE among the 3 methods.

*3.4. Backward-facing step*

Our next test case is the backward-facing step problem. As per the schematic in Figure 13, this problem describes the flow in a channel, with length and width of 20 and 1 units, respectively. The fluid from the inlet with a fully developed parabolic profile above the step flows into the channel. When the steady state is achieved, there will be a primary vortex created in the triangle region between the step and point $x_1$. As $Re$ increases, a secondary eddy will appear at the top at mid region of the channel, between $x_2$ and $x_3$. The governing equations for this test problem are the steady-state, two-dimensional incompressible N-S equations as described in Section 3.3. We solve for the problem at $Re = 200$, with the spatial domain specified as $x \in [0, 20], y \in [-0.5, 0.5]$. For comparison of our PINN models, the ground truth is simulated by the same in-house numerical solver described in Section 3.3. It has already been shown in [65] that the solutions obtained by our solver can achieve very good agreement with other benchmark results.

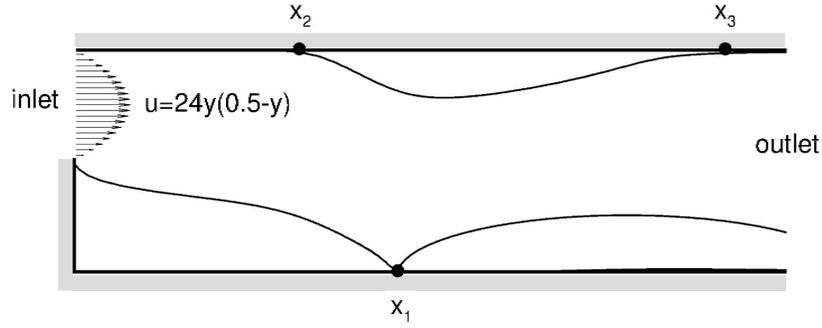

Fig. 13. Schematic of backward-facing step problem.

*3.4.1. Comparison between a-PINN, n-PINN, and can-PINNs results*

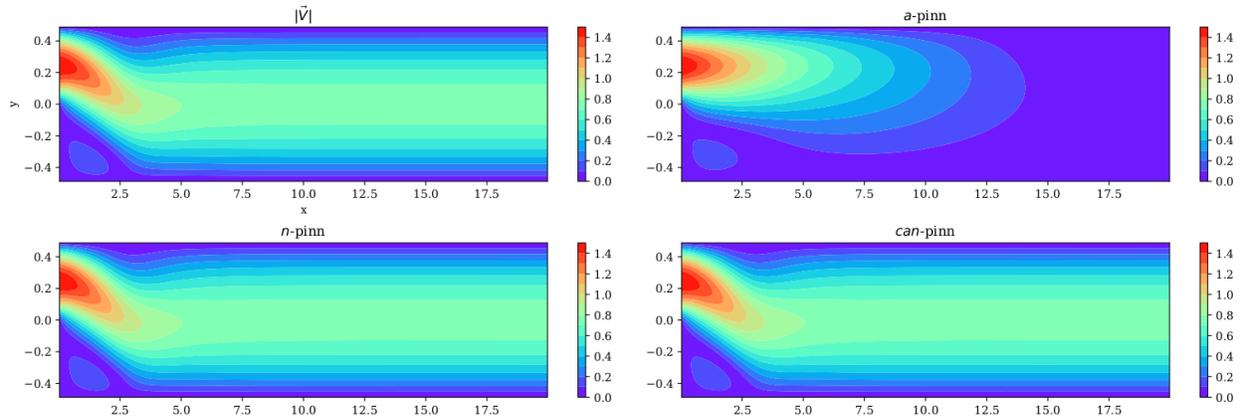

(a) Ground truth velocity magnitude $|\vec{V}|$ vs. PINN solutions

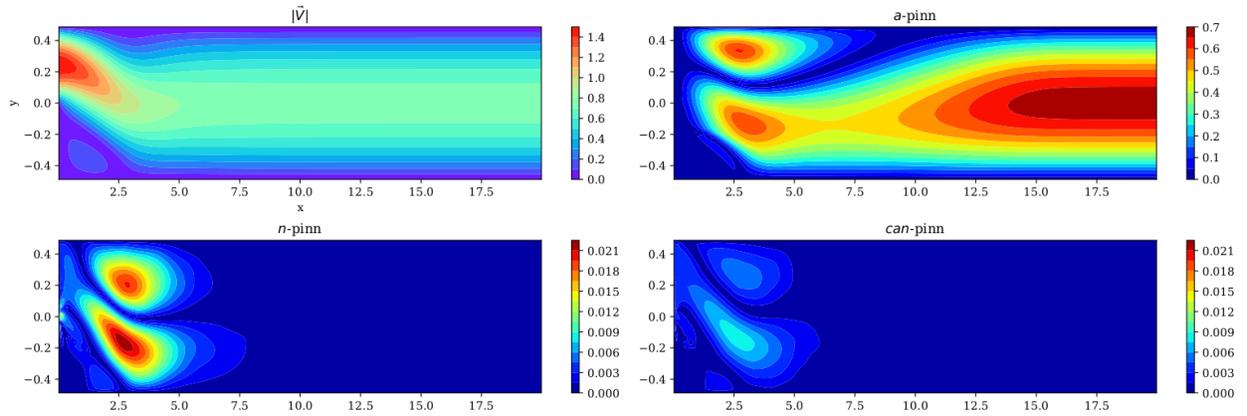

(b) Ground truth velocity magnitude $|\vec{V}|$ and absolute error of PINN solutions

Fig. 14. (a) Comparison between the ground truth velocity magnitude $|\vec{V}|$ and the solutions solved by *a*-PINN, *n*-PINN, and *can(uw2-conv, cd-p)*-PINN, for the backward facing step problem at $Re = 200$. (b) The absolute error between the PINN solutions and ground truth. From 25 independent runs, the solution with median MSE for respective PINN models are shown.

In this problem, we compare the performance of *a*-PINN, *n*-PINN, and *can*-PINNs (both *can(uw2-conv)*-PINN and *can(uw2-conv, cd-p)*-PINN), training with a set of 16,000 collocation points. For each model, 25 independent runs were performed. Figure 14 visually compares the velocity magnitude $|\vec{V}| = \sqrt{u^2 + v^2}$ contour computed from solutions solved by *a*-PINN, *n*-PINN, and *can(uw2-conv, cd-p)*-PINN, and their absolute deviation from the simulated ground truth. Moreover, the distributions of training losses and MSE against the ground truth solution are displayed in Figure 15. The median training loss are below 1e-6 for all PINN models, however the lowest training loss distribution given by *a*-PINN doesn't correspond to a more accurate solution. In fact, the *a*-PINN solutions are very poorly solved and very different from the ground truth (Figure 14) despite the low training loss. On the other hand, a sample efficient *n*-PINN consistently achieves good solutions, which are about 4 orders of magnitude lower in MSE than *a*-PINN. The proposed *can*-PINNs, in particular the *can(uw2-conv, cd-p)*-PINN consistently produces more accurate solutions. Their MSE values are 5 orders and 1 order of magnitude better than the baseline *a*-PINN and *n*-PINN, respectively. For the sake of completeness, we also tabulate the separation ($x_2$) and reattachment ($x_1$, $x_3$) points in Table 3, showing excellent agreement between our proposed *can(uw2-conv, cd-p)*-PINN and other benchmark results [68–70].

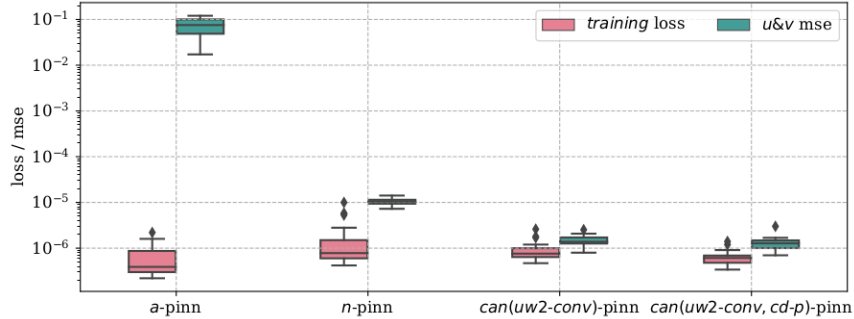

Fig. 15. Distribution of training loss and MSE between PINNs (*a*-PINN, *n*-PINN, and *can*-PINNs) and ground truth solutions for the backward facing step problem, at $Re = 200$. For $u\&v$ MSE, we compute the MSE for *u*- and *v*-velocity components and take an average. Results from 25 independent runs are shown as boxplot.

Table 3. The PINN prediction and benchmark reattachment lengths for the backward-facing step problems.

|  | $Re = 200$ | $Re = 400$ | | |
| --- | --- | --- | --- | --- |
|  | $x_1$ | $x_1$ | $x_2$ | $x_3$ |
| Barton [68] | 2.565 | 4.255 | 4.045 | 5.11 |
| Barber and Fonty [69] | 2.635 | 4.265 | 4.065 | 5.07 |
| Erturk [70] | 2.491 | 4.119 | 3.866 | 5.019 |
| **Ground truth (IDFC simulation)** | **2.608** | **4.276** | **4.093** | **5.122** |
| *a*-PINN | 4.761 | | | |
| *n*-PINN | 2.601 | | | |
| *can*-PINN | 2.604 | 4.292 | 4.120 | 5.161 |

### *3.4.2. Solving Re400 with can-PINN*

To further demonstrate the advantage of the proposed *can*-PINN framework, we move on to solve for a more challenging problem at $Re = 400$. As the problem complexity increased, a *can(uw2-conv, cd-p)*-PINN model is trained with a larger set of 32,000 collocation points for 2 million training iterations with 1000 mini-batch samples to ensure convergence. To compare the PINN solution with ground truth, we then use the trained model to predict the solution on 1600×80 grid points. The solution and its error are displayed in Figure 16, showing an excellent agreement with the simulated ground truth. The MSE for $u$-, $v$-velocity, and pressure are 8.4e-7, 6.3e-8, and 3.9e-8 respectively, suggesting that the *can*-PINN model able to remain accurate on a fine resolution prediction although it is trained with less collocation points. Remarkably, the *can*-PINN model can capture the tiny secondary eddy at the top of the channel, whereas the conventional *a*-PINN is unable to produce any solution for this problem. We also plot the streamline contours in Figure 17 and tabulate the separation and reattachment lengths in Table 3, showing excellent agreement (<1% deviation) between our proposed *can*-PINN and the ground truth.

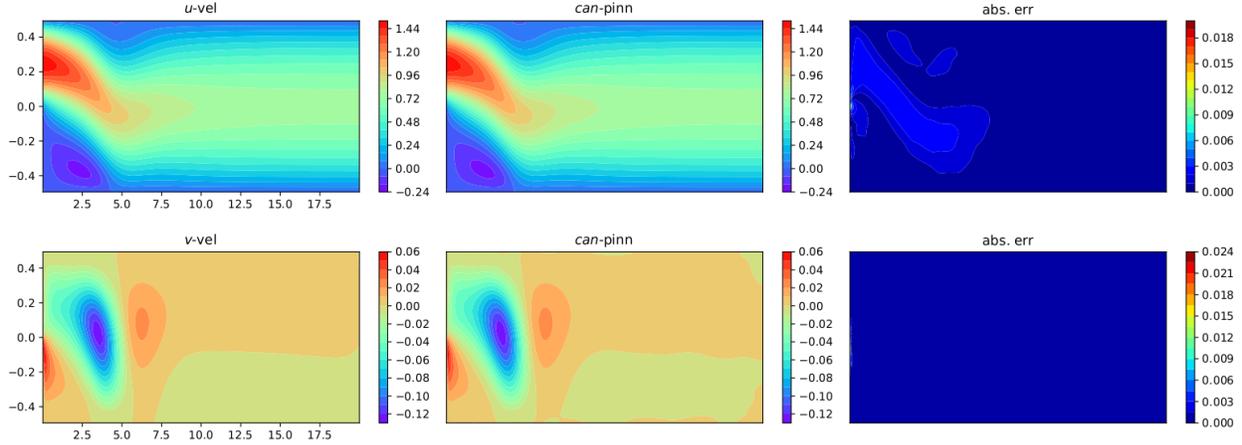

Fig. 16. The u- and v-velocity contour from the solutions obtained by IDFC simulation (1st column) and *can(uw2-conv, cd-p)*-PINN (2nd column) as well as the absolute error between *can*-PINN and simulated ground truth (3rd column), for the backward-facing step problem at $Re = 400$.

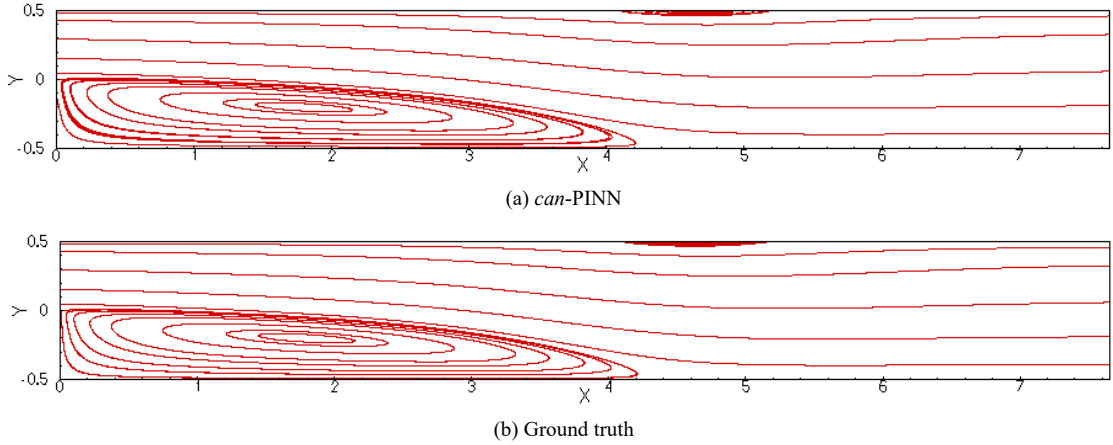

Fig. 17. The plot of streamlines generated from (a) *can(uw2-conv, cd-p)*-PINN prediction in comparison to (b) simulated ground truth on 1600x80 grid points, for the backward-facing step problem at $Re = 400$.

### 3.4.3. Solving inverse problem with can-PINN

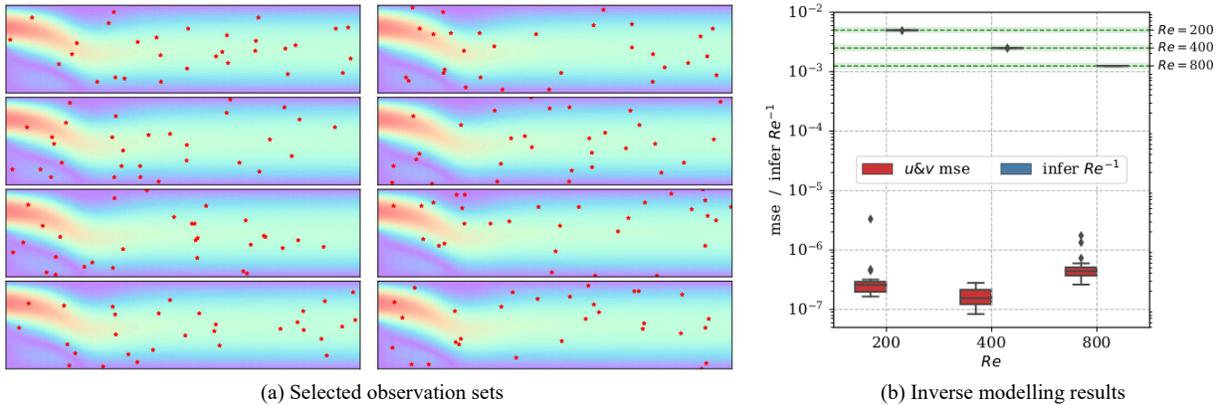

Fig. 18. (a) Eight random observation sets in the inverse backward facing step problem, where the *can(uw2-conv, cd-p)*-PINN is applied to simultaneously infer the unknown $Re$ coefficient in the incompressible N-S equations and also to solve the solution based on limited observations ($n = 30$, in red). (b) The distribution of $u\&v$ MSE and inferred $Re^{-1}$ obtained from the *can(uw2-conv, cd-p)*-PINN for $Re = 200, 400\ \&\ 800$. The green dashed lines and shaded areas indicate the ground truth $Re^{-1}$ and their 10% error bounds. Results are based on 20 independent runs.

Our final experiment is to infer the unknown $Re$ in the incompressible N-S equations (27) from observation data, as well as solving the solution for the entire domain. We test the inverse modelling capability of *can(uw2-conv, cd-p)*-PINN given separate sets of observations from the backward-facing step problem at 3 different Reynolds number, $Re = 200, 400, 800$. The models are trained with 16,000 collocation points, following the same training setting listed in Table 1. Similar to the inverse lid driven cavity problem in Section 3.3.2, the observation data only contains 30 randomly drawn samples (i.e., $u$ and $v$ values obtained from the IDFC simulations). Selected observation sets are displayed in Figure 18a. The distributions of inferred $1/Re$ and $u\&v$ MSE based on 20 independent runs are shown in Figure 18a-c. Similarly, the *can*-PINN infers the correct $Re$ ($< 2.5\%$ error) accurately for all 3 $Re$ numbers with limited observation data. All the inverse *can*-PINN solutions also achieve a $u\&v$ MSE below 5e-6. The results are also consistent across different sets of observation data, further demonstrating the effectiveness of *can*-PINNs for solving complex inverse problems with limited observations.

*3.5. Complex channel*

Our final test case is the 2D complex channel flow problem, to demonstrate the capability of the proposed *can*-PINNs on handling irregular domain. As seen in Figure 19, this problem describes the flow in a 2 units length channel. The width of the channel for inlet and outlet is 0.3 unit, while the middle section of channel has a shape of Singapore island. The problem is solved at $Re=400$ with the prescribed inlet profile $u = -\frac{y^2}{0.015} + 1.5$, non-slip BC at top/bottom wall, and freestream outlet condition. The governing equations for this test problem are the steady-state, two-dimensional incompressible N-S equations (27) as described in Section 3.3. For comparison of our PINN models, the ground truth is simulated by the same in-house numerical solver also described in Section 3.3.

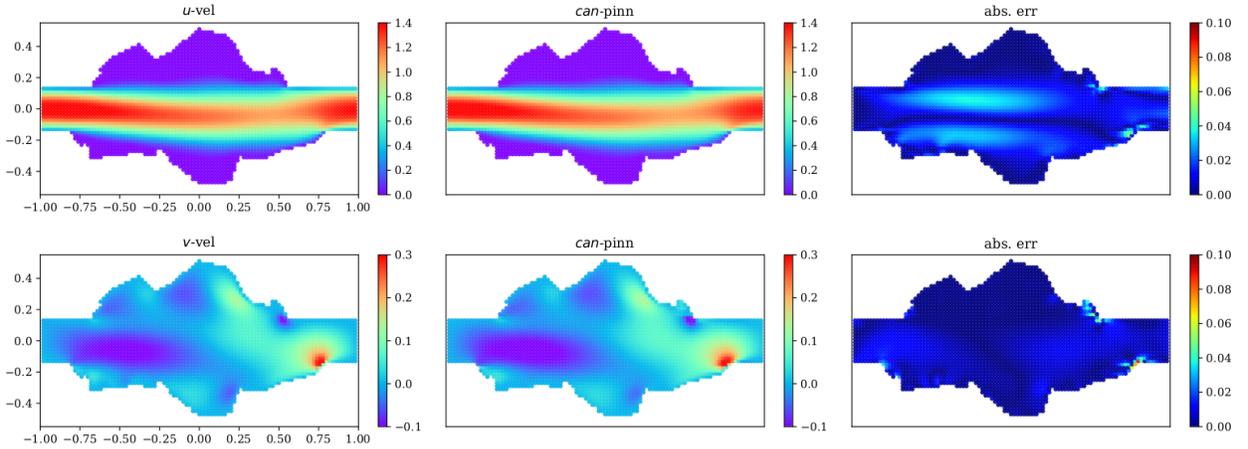

Fig. 19. The u- and v-velocity contour from the solutions obtained by IDFC simulation (1st column) and *can(uw2-conv, cd-p)*-PINN (2nd column) as well as the absolute error between *can*-PINN solution and simulated ground truth (3rd column), for the 2D complex channel problem at $Re = 400$. The medium solution from 25 independent runs is shown.

To sample the collocation points from irregular domain, our *n*-PINN and *can*-PINN methods require an additional routine to determine if the point is inside, on the boundary, or outside of the complex geometry. This requirement is the same as the *a*-PINN approach. We train *a*-PINN, *n*-PINN, and *can(uw2-conv, cd-p)*-PINN models with 674 boundary points to represent the geometry and 2,763 equidistantly spaced collocation points inside the geometry. We enforce the BCs on these 674 boundary points, and PDE loss on the 2,763 inner domain collocation points. When evaluating PDE loss for *n*-PINN and *can*-PINN, we predict the stencil values by $\hat{u}(x, y; \mathbf{w})$, whether they are in or out of the domain. The implementation is like the standard *a*-PINN. The PINN architecture and training settings are identical for all models, as listed in Table 1. For each model, we perform 25 independent runs.

*3.5.1. Comparison between a-PINN, n-PINN, and can-PINNs results*

The distribution of training losses and solution MSEs obtained from *a*-PINN, *n*-PINN, and *can*-PINN are shown in Figure 20. The results revealed that the *a*-PINN fails to produce plausible solution (i.e., large discrepancy) at this training sample density, despite having the lowest training loss. On the other hand, our present approach can achieve a good solution MSE although the problem has a complex, irregular domain.

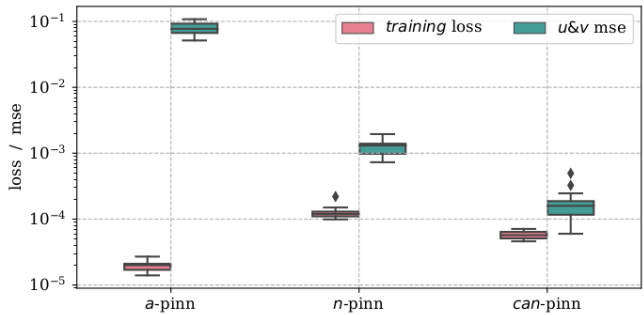

Fig. 20. Distribution of training loss and MSE between PINNs (*a*-PINNs, *n*-PINNs, and *can(uw2-conv, cd-p)*-PINNs) and ground truth solutions for the 2D complex channel problem at $Re = 400$. Results from 25 independent runs are shown as boxplot.

## 4. Conclusions

In this paper, we first studied the difference between PINNs with training loss computed by AD and our proposed ND-based approach. It was observed that the AD-formulated loss function is likely an under-constrained optimization problem, which causes the PINN training to become completely unrelated to the accuracy of its solution without sufficient sampling. We then showed that the ND-formulated PINN is much more sample efficient, with fairly good solution predictions regardless of collocation point's density.

Building on the idea of multi-moment schemes in computational physics, we further proposed a *coupled-automatic-numerical* differentiation method that utilizes both AD and values on the local support points for approximating derivative terms in the PINN loss function, unifying the advantages of both AD and ND-based approaches. The resulting *can*-PINN is not only much more sample efficient, but also yields an improved accuracy. With application to fluid dynamic problems in mind, we derived two instantiations of *can*-PINN schemes based on the upwind and central difference numerical schemes in this work. These schemes are chosen as they are critical to maintaining the convective stability and the coupling between velocity and pressure when solving the incompressible N-S equations. Fundamental analysis also revealed better dispersion and dissipation behavior for the proposed schemes, which was corroborated in our experiments.

Although demonstrated on two common numerical schemes (upwind and central difference) in this work, the proposed *can*-PINN is a generic framework that can be easily extended to many *coupled-automatic-numerical* schemes of varying form and accuracy. This makes it particularly exciting as a means of tapping upon the accumulated wealth of numerical schemes that have been developed in the realm of computational physics and scientific computing. Importantly, as observed in our own experiments and is consistent with prior work in literature, the underlying physics of the problem is critical to the selection of an appropriate differentiation scheme, and potential performance gains from being able to integrate differentiation schemes of various kinds to match the problem over a one-size-fits-all approach as commonly employed in the use of AD currently will be an interesting extension for this work. On the other hand, the use of *n*-PINN and *can*-PINN require certain numerical knowledge to choose an appropriate scheme for the problem at hand, which could potentially be a drawback relative to the *a*-PINN approach.

Also, while generally applicable to non-uniform samplings, both proposed *n*-PINNs and *can*-PINNs have a more natural implementation and corresponding choice of $\Delta x$ parameter when applied to a set of equidistantly spaced collocation points. Further development is required to understand the impact of other sampling scenarios, such as from a set of highly irregular collocation points or even a statistical distribution, on the effectiveness of the newly proposed methods, especially in scenarios where more complex geometry or physics may require more sophisticated sampling strategies.

Nonetheless, the proposed *can*-PINN showed consistently high efficiency and efficacy on all the test problems in this experimental study, out-performing conventional AD-based formulations across all settings evaluated. With the ability to efficiently train on sparse samples while robustly producing an accurate solution, the *can*-PINN formulation potentially enables the extension of the PINN methodology to even more challenging problems across a multitude of domains.

**Declaration of competing interest**

The authors declare that they have no known competing financial interests or personal relationships that could have appeared to influence the work reported in this paper.

**Acknowledgments**

This research is supported by A*STAR under its AME Programmatic programme: Explainable Physics-based AI for Engineering Modelling & Design (ePAI) [Award No. A20H5b0142].

**Reference**

[1] G.E. Karniadakis, I.G. Kevrekidis, L. Lu, P. Perdikaris, S. Wang, L. Yang, Physics-informed machine learning, Nature Reviews Physics 2021 3:6. 3 (2021) 422–440. https://doi.org/10.1038/s42254-021-00314-5.

[2] M. Raissi, P. Perdikaris, G.E. Karniadakis, Physics-informed neural networks: A deep learning framework for solving forward and inverse problems involving nonlinear partial differential equations, Journal of Computational Physics. 378 (2019) 686–707. https://doi.org/10.1016/J.JCP.2018.10.045.

[3] J.C. Wong, A. Gupta, Y.S. Ong, Can Transfer Neuroevolution Tractably Solve Your Differential Equations?, IEEE Computational Intelligence Magazine. 16 (2021) 14–30. https://doi.org/10.1109/MCI.2021.3061854.

[4] H. Lee, I.S. Kang, Neural algorithm for solving differential equations, Journal of Computational Physics. 91 (1990) 110–131.

[5] A.J. Meade, A.A. Fernandez, Solution of nonlinear ordinary differential equations by feedforward neural networks, Mathematical and Computer Modelling. 20 (1994) 19–44. https://doi.org/10.1016/0895-7177(94)00160-X.

[6] M.W.M.G. Dissanayake, N. Phan-Thien, Neural-network-based approximations for solving partial differential equations, Communications in Numerical Methods in Engineering. 10 (1994) 195–201. https://doi.org/10.1002/CNM.1640100303.

[7] B.Ph. van Milligen, V. Tribaldos, J.A. Jiménez, Neural Network Differential Equation and Plasma Equilibrium Solver, Physical Review Letters. 75 (1995) 3594. https://doi.org/10.1103/PhysRevLett.75.3594.

[8] I.E. Lagaris, A. Likas, D.I. Fotiadis, Artificial neural networks for solving ordinary and partial differential equations, IEEE Transactions on Neural Networks. 9 (1998) 987–1000. https://doi.org/10.1109/72.712178.

[9] J. Berg, K. Nyström, A unified deep artificial neural network approach to partial differential equations in complex geometries, Neurocomputing. 317 (2018) 28–41.

[10] E. Haghighat, M. Raissi, A. Moure, H. Gomez, R. Juanes, A physics-informed deep learning framework for inversion and surrogate modeling in solid mechanics, Computer Methods in Applied Mechanics and Engineering. 379 (2021) 113741. https://doi.org/10.1016/J.CMA.2021.113741.

[11] M.A. Nabian, H. Meidani, Physics-Driven Regularization of Deep Neural Networks for Enhanced Engineering Design and Analysis, Journal of Computing and Information Science in Engineering. 20 (2020). https://doi.org/10.1115/1.4044507.

[12] Z. Fang, A High-Efficient Hybrid Physics-Informed Neural Networks Based on Convolutional Neural Network, IEEE Transactions on Neural Networks and Learning Systems. (2021). https://doi.org/10.1109/TNNLS.2021.3070878.

[13] R. Ranade, C. Hill, J. Pathak, DiscretizationNet: A machine-learning based solver for Navier–Stokes equations using finite volume discretization, Computer Methods in Applied Mechanics and Engineering. 378 (2021) 113722. https://doi.org/10.1016/J.CMA.2021.113722.

[14] N. Wandel, M. Weinmann, R. Klein, Learning Incompressible Fluid Dynamics from Scratch - Towards Fast, Differentiable Fluid Models that Generalize, Proceedings of International Conference on Learning Representations (ICLR). (2021). https://arxiv.org/abs/2006.08762 (accessed August 14, 2021).

[15] N. Wandel, M. Weinmann, R. Klein, Teaching the incompressible Navier–Stokes equations to fast neural surrogate models in three dimensions, Physics of Fluids. 33 (2021) 047117. https://doi.org/10.1063/5.0047428.

[16] H. Gao, L. Sun, J.X. Wang, PhyGeoNet: Physics-informed geometry-adaptive convolutional neural networks for solving parameterized steady-state PDEs on irregular domain, Journal of Computational Physics. 428 (2021) 110079. https://doi.org/10.1016/J.JCP.2020.110079.

[17] J. Sirignano, K. Spiliopoulos, DGM: A deep learning algorithm for solving partial differential equations, Journal of Computational Physics. 375 (2018) 1339–1364.

[18] N. Geneva, N. Zabaras, Modeling the dynamics of PDE systems with physics-constrained deep auto-regressive networks, Journal of Computational Physics. 403 (2020) 109056. https://doi.org/10.1016/J.JCP.2019.109056.

[19] R. Zhang, Y. Liu, H. Sun, Physics-informed multi-LSTM networks for metamodeling of nonlinear structures, Computer Methods in Applied Mechanics and Engineering. 369 (2020) 113226. https://doi.org/10.1016/J.CMA.2020.113226.

[20] P. Ren, C. Rao, Y. Liu, J. Wang, H. Sun, PhyCRNet: Physics-informed Convolutional-Recurrent Network for Solving Spatiotemporal PDEs, (2021). https://arxiv.org/abs/2106.14103v1 (accessed August 14, 2021).

[21] Y. Yang, P. Perdikaris, Adversarial uncertainty quantification in physics-informed neural networks, Journal of Computational Physics. 394 (2019) 136–152. https://doi.org/10.1016/J.JCP.2019.05.027.

[22] L. Yang, D. Zhang, G.E. Karniadakis, Physics-Informed Generative Adversarial Networks for Stochastic Differential Equations, SIAM Journal on Scientific Computing. 42 (2020) A292–A317. https://doi.org/10.1137/18M1225409.

[23] N. Zobeiry, K.D. Humfeld, A physics-informed machine learning approach for solving heat transfer equation in advanced manufacturing and engineering applications, Engineering Applications of Artificial Intelligence. 101 (2021) 104232. https://doi.org/10.1016/J.ENGAPPAI.2021.104232.

[24] S. Amini Niaki, E. Haghighat, T. Campbell, A. Poursartip, R. Vaziri, Physics-informed neural network for modelling the thermochemical curing process of composite-tool systems during manufacture, Computer Methods in Applied Mechanics and Engineering. 384 (2021). https://doi.org/10.1016/J.CMA.2021.113959.

[25] X. Jin, S. Cai, H. Li, G.E. Karniadakis, NSFnets (Navier-Stokes flow nets): Physics-informed neural networks for the incompressible Navier-Stokes equations, Journal of Computational Physics. 426 (2021) 109951. https://doi.org/10.1016/J.JCP.2020.109951.

[26] L. Sun, H. Gao, S. Pan, J.X. Wang, Surrogate modeling for fluid flows based on physics-constrained deep learning without simulation data, Computer Methods in Applied Mechanics and Engineering. 361 (2020).

[27] Z. Mao, A.D. Jagtap, G.E. Karniadakis, Physics-informed neural networks for high-speed flows, Computer Methods in Applied Mechanics and Engineering. 360 (2020).

[28] Z. Fang, J. Zhan, Deep Physical Informed Neural Networks for Metamaterial Design, IEEE Access. 8 (2020) 24506–24513. https://doi.org/10.1109/ACCESS.2019.2963375.

[29] P. Zhang, Y. Hu, Y. Jin, S. Deng, X. Wu, J. Chen, A maxwell's equations based deep learning method for time domain electromagnetic simulations, Proceedings of the 2020 IEEE Texas Symposium on Wireless and Microwave Circuits and Systems: Making Waves in Texas, WMCS 2020. (2020). https://doi.org/10.1109/WMCS49442.2020.9172407.

[30] I.E. Lagaris, A. Likas, D.I. Fotiadis, Artificial neural network methods in quantum mechanics, Computer Physics Communications. 104 (1997) 1–14. https://doi.org/10.1016/S0010-4655(97)00054-4.

[31] M. Raissi, Deep Hidden Physics Models: Deep Learning of Nonlinear Partial Differential Equations, Journal of Machine Learning Research. 19 (2018) 1–24.

[32] M. Raissi, A. Yazdani, G.E. Karniadakis, Hidden fluid mechanics: Learning velocity and pressure fields from flow visualizations, Science. 367 (2020) 1026–1030. https://doi.org/10.1126/SCIENCE.AAW4741.

[33] G. Kissas, Y. Yang, E. Hwuang, W.R. Witschey, J.A. Detre, P. Perdikaris, Machine learning in cardiovascular flows modeling: Predicting arterial blood pressure from non-invasive 4D flow MRI data using physics-informed neural networks, Computer Methods in Applied Mechanics and Engineering. 358 (2020).

[34] M. Raissi, Z. Wang, M.S. Triantafyllou, G.E. Karniadakis, Deep learning of vortex-induced vibrations, Journal of Fluid Mechanics. 861 (2019) 119–137. https://doi.org/10.1017/JFM.2018.872.

[35] Y. Chen, L. Lu, G.E. Karniadakis, L. Dal Negro, Physics-informed neural networks for inverse problems in nano-optics and metamaterials, Optics Express. 28 (2020) 11618. https://doi.org/10.1364/OE.384875.

[36] K. Shukla, P.C. Di Leoni, J. Blackshire, D. Sparkman, G.E. Karniadakis, Physics-Informed Neural Network for Ultrasound Nondestructive Quantification of Surface Breaking Cracks, Journal of Nondestructive Evaluation 2020 39:3. 39 (2020) 1–20. https://doi.org/10.1007/S10921-020-00705-1.

[37] J.C. Wong, C. Ooi, A. Gupta, Y.-S. Ong, Learning in Sinusoidal Spaces with Physics-Informed Neural Networks, (2021).

[38] S. Wang, H. Wang, P. Perdikaris, On the eigenvector bias of Fourier feature networks: From regression to solving multi-scale PDEs with physics-informed neural networks, Computer Methods in Applied Mechanics and Engineering. 384 (2021) 113938. https://doi.org/10.1016/J.CMA.2021.113938.

[39] R. van der Meer, C. Oosterlee, A. Borovykh, Optimally weighted loss functions for solving PDEs with Neural Networks, (2020).

[40] S. Wang, Y. Teng, P. Perdikaris, Understanding and mitigating gradient pathologies in physics-informed neural networks, (2020).

[41] S. Wang, X. Yu, P. Perdikaris, When and why PINNs fail to train: A neural tangent kernel perspective, (2020).

[42] L. McClenny, U. Braga-Neto, Self-Adaptive Physics-Informed Neural Networks using a Soft Attention Mechanism, (2020).

[43] M.A. Nabian, R.J. Gladstone, H. Meidani, Efficient training of physics-informed neural networks via importance sampling, Computer-Aided Civil and Infrastructure Engineering. 36 (2021) 962–977. https://doi.org/10.1111/MICE.12685.

[44] O. Fuks, H.A. Tchelepi, LIMITATIONS OF PHYSICS INFORMED MACHINE LEARNING FOR NONLINEAR TWO-PHASE TRANSPORT IN POROUS MEDIA, Journal of Machine Learning for Modeling and Computing. 1 (2020) 19–37. https://doi.org/10.1615/.2020033905.

[45] C.F. Gasmi, H. Tchelepi, Physics Informed Deep Learning for Flow and Transport in Porous Media, (2021).

[46] O. Hennigh, S. Narasimhan, M.A. Nabian, A. Subramaniam, K. Tangsali, M. Rietmann, J. del A. Ferrandis, W. Byeon, Z. Fang, S. Choudhry, NVIDIA SimNet^{TM}: an AI-accelerated multi-physics simulation framework, (2020) 447–461.

[47] E. Haghighat, R. Juanes, SciANN: A Keras/TensorFlow wrapper for scientific computations and physics-informed deep learning using artificial neural networks, Computer Methods in Applied Mechanics and Engineering. 373 (2021) 113552. https://doi.org/10.1016/J.CMA.2020.113552.

[48] L. Lu, X. Meng, Z. Mao, G.E. Karniadakis, DeepXDE: A Deep Learning Library for Solving Differential Equations, SIAM Review. 63 (2021) 208–228. https://doi.org/10.1137/19M1274067.

[49] A. Güne¸, G. Baydin, B.A. Pearlmutter, J.M. Siskind, Automatic Differentiation in Machine Learning: a Survey, Journal of Machine Learning Research. 18 (2018) 1–43.

[50] D. Anderson, J.C. Tannehill, R.H. Pletcher, Computational fluid mechanics and heat transfer, Third edition, CRC Press, Fourth edition. | Boca Raton, FL : CRC Press, 2020. | Series: Computational and physical processes in mechanics and thermal sciences, 2016. https://doi.org/10.1201/9781351124027.

[51] H. Gao, M.J. Zahr, J.-X. Wang, Physics-informed graph neural Galerkin networks: A unified framework for solving PDE-governed forward and inverse problems, Computer Methods in Applied Mechanics and Engineering. 390 (2022) 114502. https://doi.org/10.1016/J.CMA.2021.114502.

[52] E. Haghighat, A.C. Bekar, E. Madenci, R. Juanes, A nonlocal physics-informed deep learning framework using the peridynamic differential operator, Computer Methods in Applied Mechanics and Engineering. 385 (2021) 114012. https://doi.org/10.1016/J.CMA.2021.114012.

[53] Y. Shin, J. Darbon, G.E. Karniadakis, On the convergence of physics informed neural networks for linear second-order elliptic and parabolic type PDEs, Communications in Computational Physics. 28 (2020) 2042–2074. https://doi.org/10.4208/CICP.OA-2020-0193.

[54] D.P. Kingma, J.L. Ba, Adam: A method for stochastic optimization, 3rd International Conference on Learning Representations, ICLR 2015 - Conference Track Proceedings. (2015).

[55] K. He, X. Zhang, S. Ren, J. Sun, Delving Deep into Rectifiers: Surpassing Human-Level Performance on ImageNet Classification, in: 2015 IEEE International Conference on Computer Vision (ICCV), IEEE, 2015: pp. 1026–1034. https://doi.org/10.1109/ICCV.2015.123.

[56] V. Sitzmann, J. Martel, A. Bergman, D. Lindell, G. Wetzstein, Implicit Neural Representations with Periodic Activation Functions, Advances in Neural Information Processing Systems. 33 (2020) 7462–7473.

[57]   Y. Bengio, Practical Recommendations for Gradient-Based Training of Deep Architectures, Lecture Notes in Computer Science (Including Subseries Lecture Notes in Artificial Intelligence and Lecture Notes in Bioinformatics). 7700 LECTURE NO (2012) 437–478. https://doi.org/10.1007/978-3-642-35289-8_26.

[58]   XiaoFeng, Unified formulation for compressible and incompressible flows by using multi-integrated moments I, Journal of Computational Physics. 195 (2004) 629–654. https://doi.org/10.1016/J.JCP.2003.10.014.

[59]   K. Yokoi, M. Furuichi, M. Sakai, An efficient multi-dimensional implementation of VSIAM3 and its applications to free surface flows, Physics of Fluids. 29 (2017) 121611. https://doi.org/10.1063/1.4996183.

[60]   P.H. Chiu, H.J. Poh, Development of an improved divergence-free-condition compensated coupled framework to solve flow problems with time-varying geometries, International Journal for Numerical Methods in Fluids. 93 (2021) 44–70. https://doi.org/10.1002/fld.4874.

[61]   T.W.H. Sheu, P.H. Chiu, A divergence-free-condition compensated method for incompressible Navier–Stokes equations, Computer Methods in Applied Mechanics and Engineering. 196 (2007) 4479–4494. https://doi.org/10.1016/J.CMA.2007.05.015.

[62]   P.H. Chiu, T.W.H. Sheu, R.K. Lin, An effective explicit pressure gradient scheme implemented in the two-level non-staggered grids for incompressible Navier-Stokes equations, Journal of Computational Physics. 227 (2008) 4018–4037. https://doi.org/10.1016/j.jcp.2007.12.007.

[63]   P. Tamamidis, D.N. Assanis, Evaluation of various high-order-accuracy schemes with and without flux limiters, International Journal for Numerical Methods in Fluids. 16 (1993) 931–948. https://doi.org/10.1002/FLD.1650161006.

[64]   E. Erturk, T.C. Corke, C. Gökçöl, Numerical solutions of 2-D steady incompressible driven cavity flow at high Reynolds numbers, International Journal for Numerical Methods in Fluids. 48 (2005) 747–774. https://doi.org/10.1002/FLD.953.

[65]   P.H. Chiu, An improved divergence-free-condition compensated method for solving incompressible flows on collocated grids, Computers and Fluids. 162 (2018) 39–54. https://doi.org/10.1016/j.compfluid.2017.12.005.

[66]   U. Ghia, K.N. Ghia, C.T. Shin, High-Re solutions for incompressible flow using the Navier-Stokes equations and a multigrid method, Journal of Computational Physics. 48 (1982) 387–411. https://doi.org/10.1016/0021-9991(82)90058-4.

[67]   M. Abadi, P. Barham, J. Chen, Z. Chen, A. Davis, J. Dean, M. Devin, S. Ghemawat, G. Irving, M. Isard, M. Kudlur, J. Levenberg, R. Monga, S. Moore, D.G. Murray, B. Steiner, P. Tucker, V. Vasudevan, P. Warden, M. Wicke, Y. Yu, X. Zheng, TensorFlow: A system for large-scale machine learning, Proceedings of the 12th USENIX Symposium on Operating Systems Design and Implementation, OSDI 2016. (2016) 265–283. https://arxiv.org/abs/1605.08695v2 (accessed August 13, 2021).

[68]   I.E. Barton, A numerical study of flow over a confined backward-facing step, International Journal for Numerical Methods in Fluids. 21 (1995) 653–665. https://doi.org/10.1002/FLD.1650210804.

[69]   R.W. Barber, A. Fonty, Numerical Simulation Of Confined Laminar Flow Over A Backward- Facing Step Using A Novel Viscous-splitting Vortex Algorithm, WIT Transactions on Modelling and Simulation. 30 (2001) 1018. https://doi.org/10.2495/CMEM010111.

[70]   E. Erturk, Numerical solutions of 2-D steady incompressible flow over a backward-facing step, Part I: High Reynolds number solutions, Computers and Fluids. 37 (2008) 633–655. https://doi.org/10.1016/j.compfluid.2007.09.003.